\documentclass[journal=jacsat,manuscript=article]{achemso}

\usepackage[version=3]{mhchem} % Formula subscripts using \ce{}
\usepackage{siunitx}
\usepackage{subcaption}
\usepackage{graphicx}
\usepackage{bbm}
\usepackage{amsfonts}
\usepackage{xcolor}
\usepackage{lineno}
\author{Yuyang Wang}
\affiliation[meche]
{Department of Mechanical Engineering, Carnegie Mellon University, PA 15213, USA}
\altaffiliation{Joint First Authorship}
\author{Zhonglin Cao}
\affiliation[meche]
{Department of Mechanical Engineering, Carnegie Mellon University, PA 15213, USA}
\altaffiliation{Joint First Authorship}
\author{Amir Barati Farimani}
\email{barati@cmu.edu}
\affiliation[meche]
{Department of Mechanical Engineering, Carnegie Mellon University, PA 15213, USA}
\alsoaffiliation[cheme]
{Department of Chemical Engineering, Carnegie Mellon University, PA 15213, USA}
\title{Deep Reinforcement Learning Optimizes Graphene Nanopores for Efficient Desalination}
\abbreviations{IR,NMR,UV}

\begin{document}

\begin{abstract}
% \singlespacing 
\onehalfspacing 
Two-dimensional nanomaterials, such as graphene, have been extensively studied because of their outstanding physical properties. Structure and geometry optimization of nanopores on such materials is beneficial for their performances in real-world engineering applications, like water desalination. However, the optimization process often involves very large number of experiments or simulations which are expensive and time-consuming. In this work, we propose a graphene nanopore optimization framework via the combination of deep reinforcement learning (DRL) and convolutional neural network (CNN) for efficient water desalination. The DRL agent controls the growth of nanopore by determining the atom to be removed at each timestep, while the CNN predicts the performance of nanoporus graphene for water desalination: the water flux and ion rejection at a certain external pressure. With the synchronous feedback from CNN-accelerated desalination performance prediction, our DRL agent can optimize the nanoporous graphene efficiently in an online manner. Molecular dynamics (MD) simulations on promising DRL-designed graphene nanopores show that they have higher water flux while maintaining rival ion rejection rate compared to the normal circular nanopores. Semi-oval shape with rough edges geometry of DRL-designed pores is found to be the key factor for their high water desalination performance. Ultimately, this study shows that DRL can be a powerful tool for material design. 
\end{abstract}

\section{Introduction}

Single-layer graphene, as an iconic two-dimensional (2D) material, has drawn much scientific attention in recent decades. Because of its ultrathin thickness and outstanding mechanical properties, graphene with artificial pores has been demonstrated to have great potentials in many engineering applications such as effective hydrogen gas separator\cite{jiang2009porous,li2013ultrathin,kim2013selective}, next-generation energy storage or supercapacitor building\cite{wang2009supercapacitor,liu2010graphene}, and high-resolution DNA sequencing\cite{farimani2014dna, barati2017dna, schneider2010dna}. Given the potential imminent global water scarcity crisis, another important application for nanoporus graphene is energy-efficient water desalination\cite{cohen2012water,surwade2015water}. Equipped with nanoporous 2D material membranes like graphene, reverse osmosis (RO) water desalination process can expect 2-3 orders improvement in water flux compared with traditional polymeric membranes\cite{cohen2012water, surwade2015water,heiranian2015water,cao2019water,cao2020single}. In RO, the geometry of nanopores in 2D materials plays a determinant role in water desalination performance\cite{cohen2012water,heiranian2015water}. In general, a large pore that allows high water flux is likely to perform poorly in rejecting ions; a small pore that rejects 100\% undesired ions, on the other hand, usually have limited water flux. Thus, an optimal nanopore for water desalination is expected to allow as high water flux as possible while maintaining a high ion rejection rate. However, finding the optimal nanopore geometry on graphene can be challenging due to high computational and experimental cost associated with extensive experiments, i.e., there are millions of possible shapes for a pore on a 4nm $\times$ 4nm graphene membrane, but evaluating the water flux and ion rejection of a single pore using 10ns MD simulation takes roughly 36 hours on a 56-core CPU cluster.
% millions of pores with different shapes can be created on a 4nm $\times$ 4nm graphene membrane and 10ns of molecular dynamics (MD) simulation, which runs for 36 hours on 56 CPU cores, are necessary to evaluate the water flux/ion rejection rate of a single pore. 
Given this time benchmark, evaluating the water desalination performance of 1000 graphene nanopores can take more than 4 years. 
Therefore, to design the optimal nanopore geometry for water desalination, an efficient nanopore optimization method and a fast nanopore water desalination performance predictor (performance predictor in short) are needed. Inspired by the recent success of deep learning \cite{lecun2015deep} and reinforcement learning \cite{mnih2015human}, we combined the state-of-art deep reinforcement learning (DRL) algorithm with convolutional neural network (CNN) to solve this challenge. 

The main idea of reinforcement learning (RL)\cite{sutton1998introduction} is to train an agent to find an optimal policy which maximizes the expected return in the future through actively interacting with the environment to achieve a goal. Recently, DRL\cite{mnih2013playing,mnih2015human}, which models the RL agent with artificial neural networks, has proven to be an efficient tool in material-related engineering fields, such as material design\cite{popova2018deep, karamad2020orbital, yao2020inverse} and molecule optimization\cite{zhou2019optimization}. In this work, we designed and implemented an artificial intelligence framework consisting of DRL, which is capable of designing the nanopore on a single-layer graphene membrane to reach optimal water desalination performance. By a series of decisions on whether or not to remove carbon atoms and which atom to be removed, the DRL agent can eventually create a pore that allows the highest water flux while maintaining ion rejection rate above an acceptable threshold. Such precisely controlled atom-by-atom removal nanopore synthesis can be conducted by electrochemical reaction (ECR) \cite{russo2012atom, feng2015electrochemical}. During training, the DRL agent updates the neural network weights through back-propagation from the reward given by the water desalination performance. In our case, the evaluation of desalination performance is the water flux and ion rejection rate of the nanoporus graphene generated by the DRL agent at each timestep. However, conventional methods to calculate desalination performance, like MD simulation, are too time-consuming to be implemented in our DRL model. 
% Currently, the conventional way to obtain the water flux and ion rejection of a porous graphene membrane is through MD simulations. However, 10 ns of MD simulation can take up to several days to run, thus rendering the feedback process too time-consuming for DRL to be practical. 
To evaluate DRL-designed nanopores fast and accurately, we implemented a CNN-based \cite{lecun1998gradient,krizhevsky2012imagenet,simonyan2014very,he2016deep} model that uses the geometry of porous graphene membrane to directly predict the water flux and ion rejection rate under certain external pressure. To this end, a ResNet \cite{he2016deep} model is trained on the dataset we collected through MD simulation of water desalination using various nanoporus graphenes.  With the CNN-accelerated desalination performance prediction, the DRL model can rapidly optimize graphene nanopore for water desalination. MD simulations on promising DRL-designed nanoporus graphenes prove that they have higher water flux while maintaining similar ion rejection rate comparing to the circular nanopores. Further investigation of molecular trajectories illustrates why DRL-designed nanopores outperform the conventional nano-structures and provides insights for optimal nanopore design for water desalination.

\section{Method}

\begin{figure}[h]
    \centering
    \includegraphics[width=\textwidth, keepaspectratio=true]{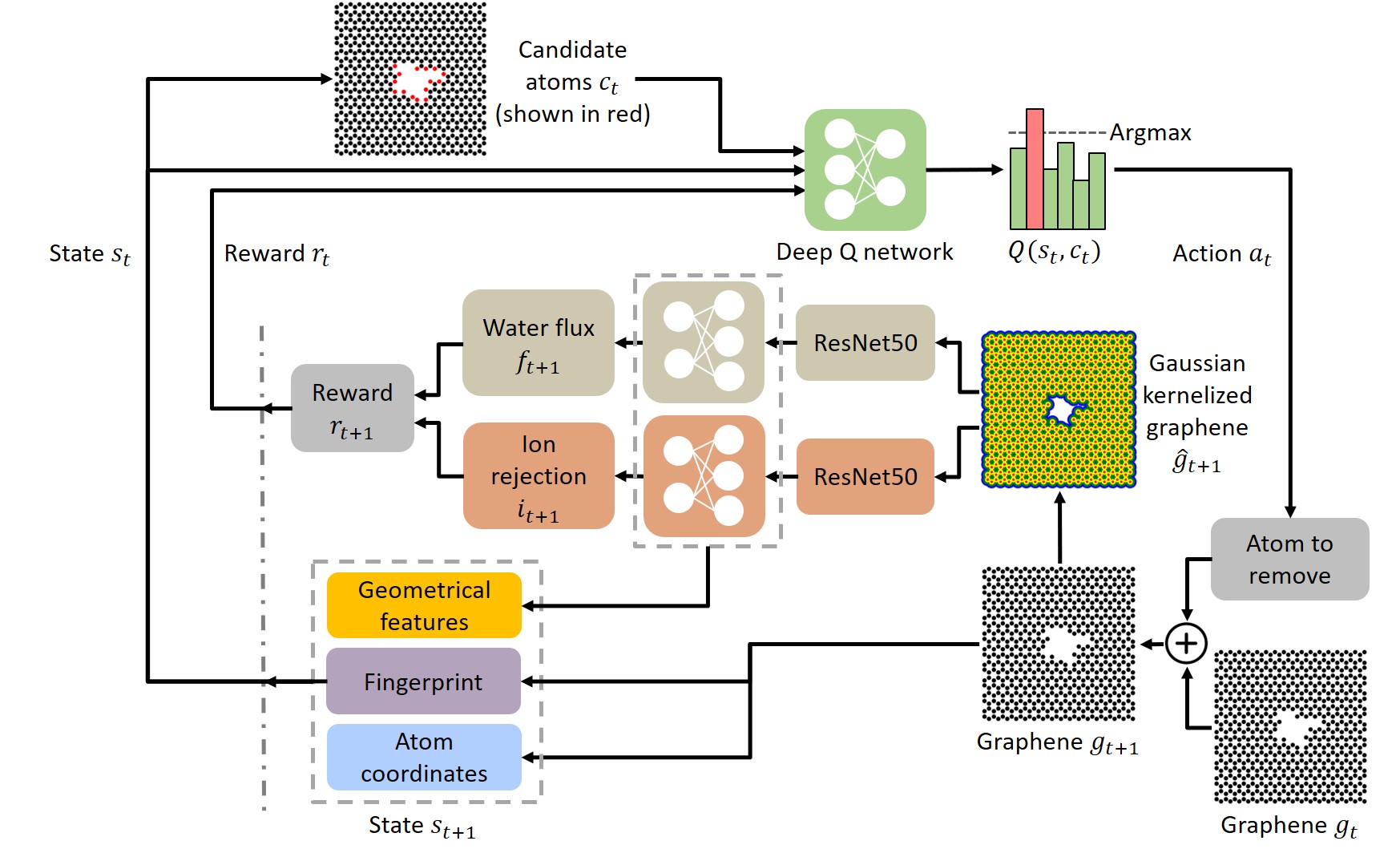}
    \caption{Overview of CNN accelerated DRL nanopore design model. At each timestep, the nanoporous graphene structure is transformed into geometrical features, which is fed into a performance prediction network to predict water flux and ion rejection rate. The reward is then calculated based on the predicted water flux and ion rejection. Also, the geometrical features extracted from the performance predictor are concatenated with the fingerprint and atom coordinates for the current state. Given the current graphene structure, candidate atoms are picked from those locate at the edge of the nanopore. The DRL agent constructed upon Deep Q-network takes the reward, candidate atoms, and state as input to determine the next atom to remove from the graphene. }
    \label{fig:pipeline}
\end{figure}

The graphene nanopore optimization framework for water desalination consists of a DRL agent incorporated with a CNN-based water desalination performance predictor (Fig.~\ref{fig:pipeline}). At each timestep, the DRL agent generates a new nanopore by removing one atom from the graphene sheet, and the CNN model poredicts the water flux/ion rejection rate of the nanopore, such that the DRL agent can get instantaneous feedback on its action. Given the featurized information of the nanoporous graphene sheet (Morgan fingerprint, Cartesian coordinates of each atom, and geometrical features of graphene membrane from the CNN model) and predicted water flux and ion rejection, the DRL agent was trained to create a pore on graphene sheet with the goal to maximize its performance in the water desalination process. The dataset used to train CNN performance predictor is generated by MD simulations of various nanoporus graphenes for water desalination.

\subsection{Molecular Dynamic Simulation}
\begin{figure}[h]
    \centering
    \begin{subfigure}[b]{0.7\textwidth}
        \centering
        \includegraphics[width=1.1\textwidth, keepaspectratio=true]{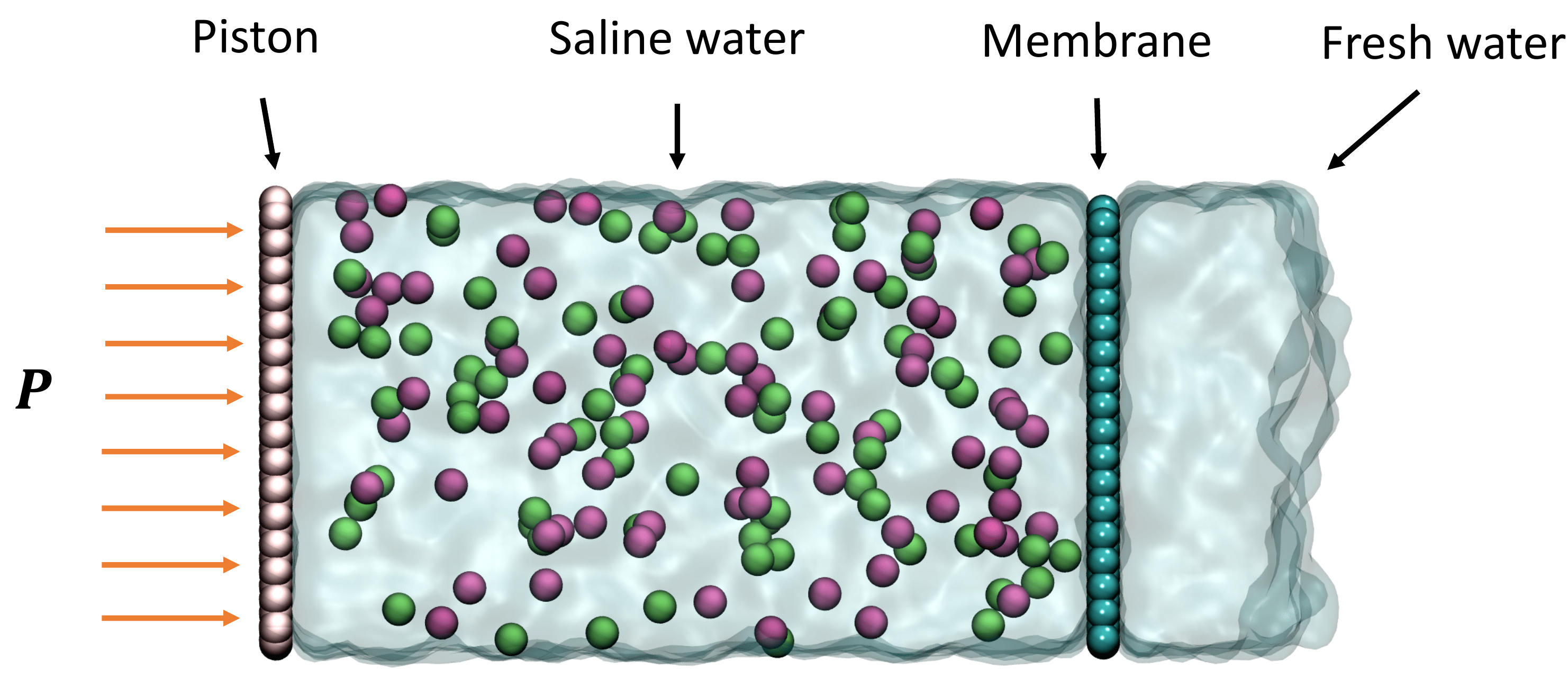}
        \caption{}
        \label{subfig:MD_system}
    \end{subfigure}
            
    \begin{subfigure}[b]{0.3\textwidth}
        \centering
        \includegraphics[width=1.1\textwidth, keepaspectratio=true]{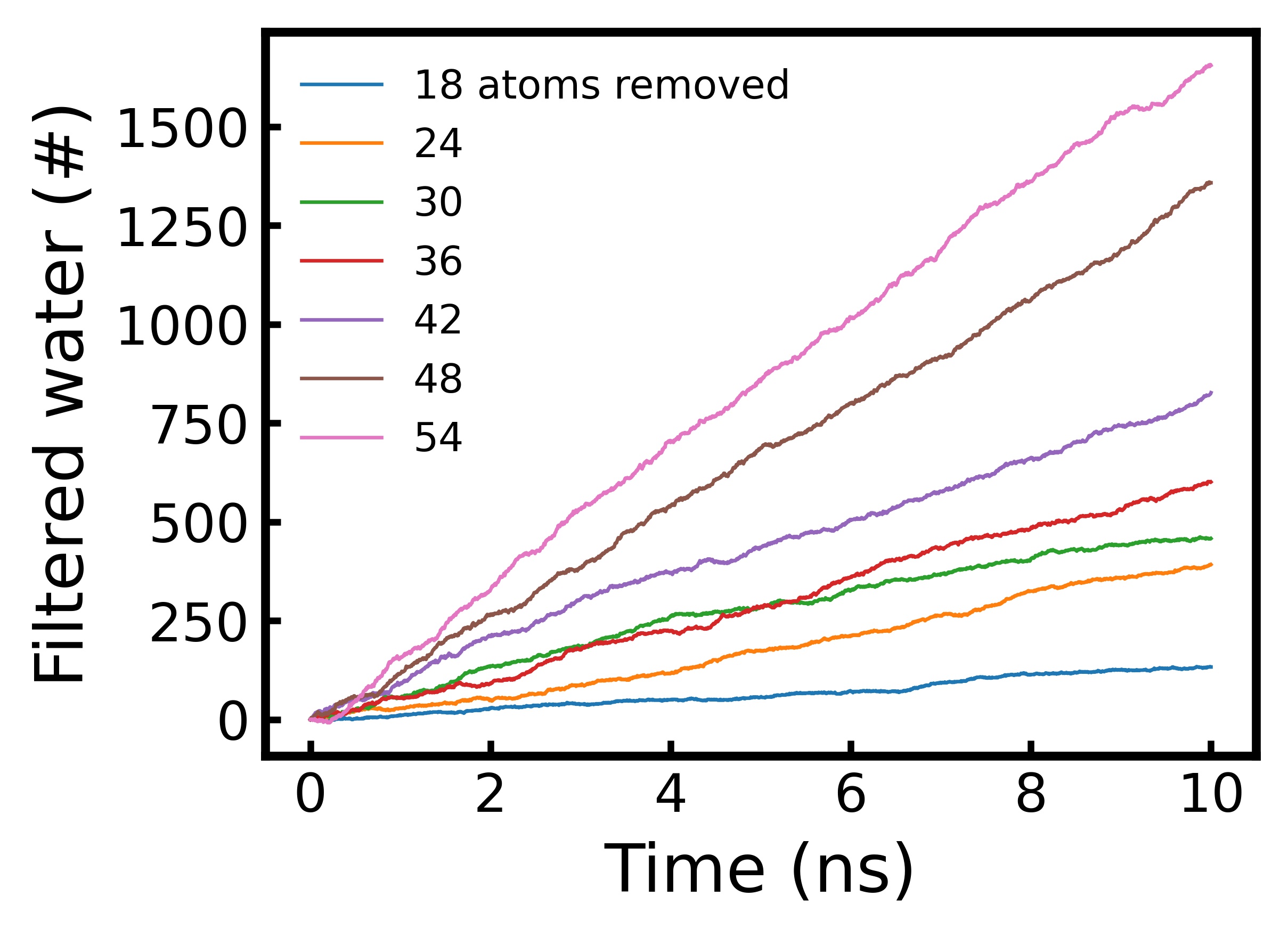}
        \caption{}
        \label{subfig:water_time}
    \end{subfigure}
    \hfill
    \begin{subfigure}[b]{0.3\textwidth}
        \centering
        \includegraphics[width=1.1\textwidth, keepaspectratio=true]{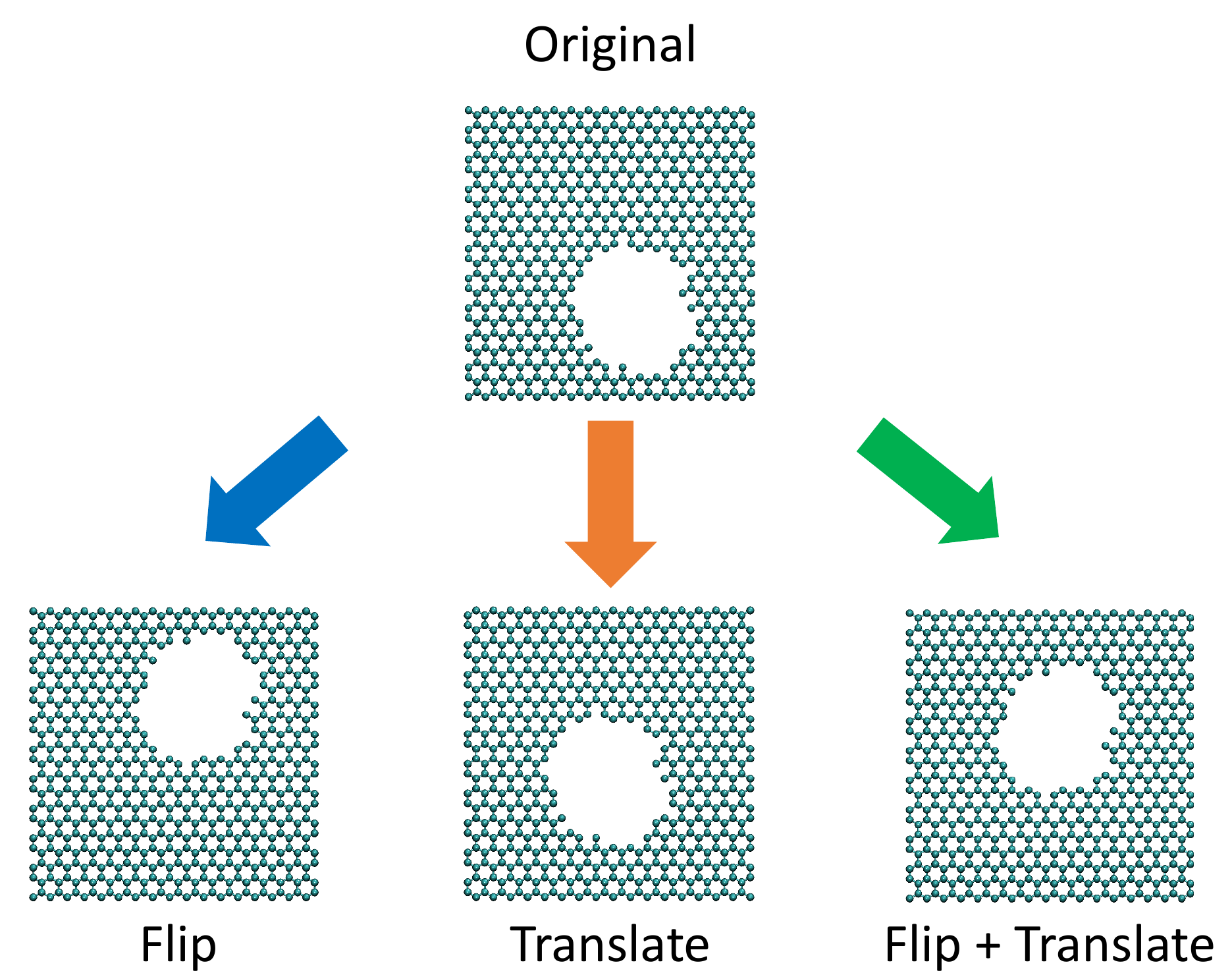}
        \caption{}
        \label{subfig:augmentation}
    \end{subfigure}
    \hfill
    \begin{subfigure}[b]{0.3\textwidth}
        \centering
        \includegraphics[width=1.1\textwidth, keepaspectratio=true]{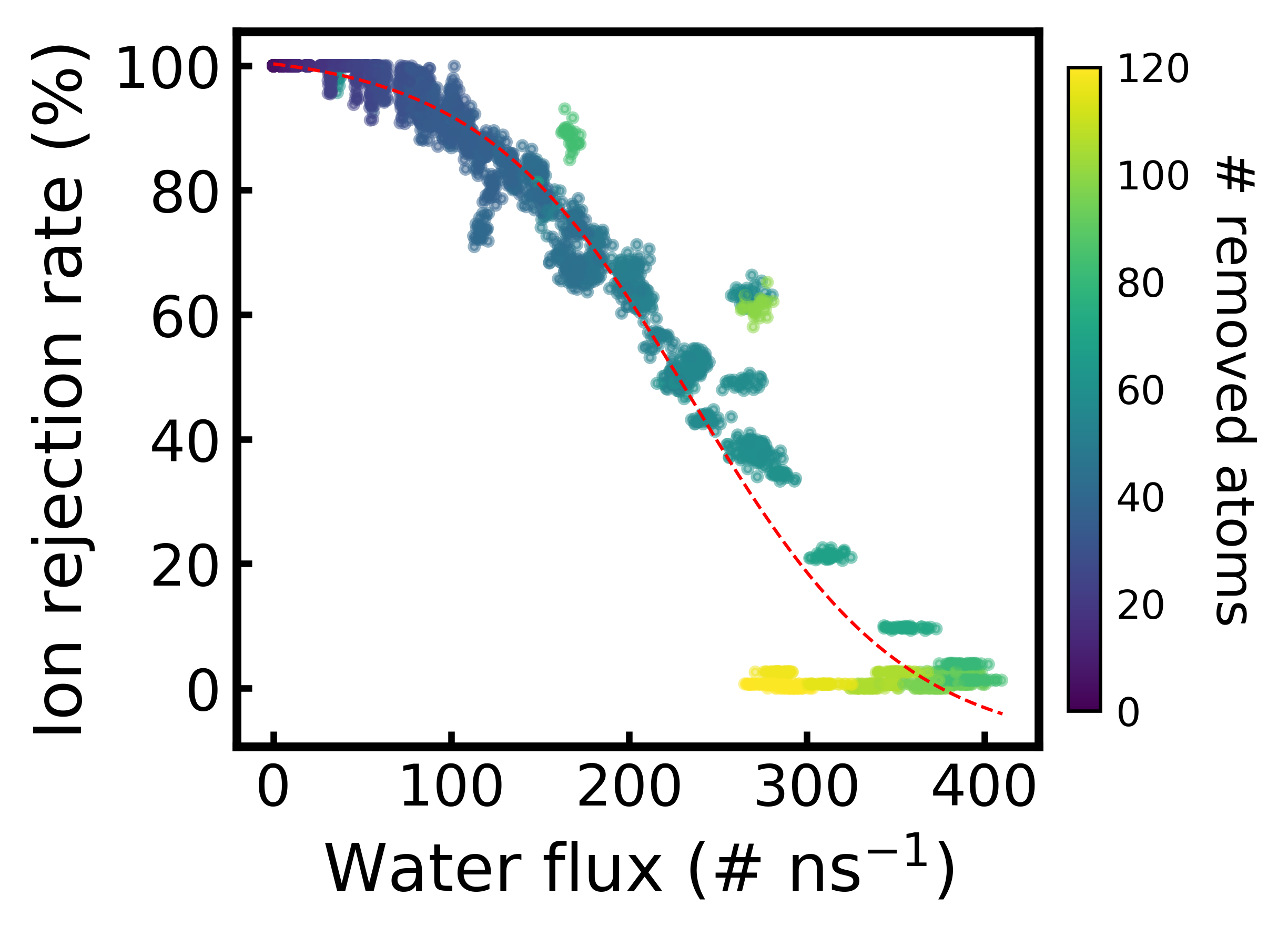}
        \caption{}
        \label{subfig:flux_rej}
    \end{subfigure}
    \hfill
    \caption{Dataset generation using MD simulation and data processing. (a) Graphene nanoporous membrane water desalination system in MD simulation. (b) Number of filtered water molecules with respect to simulation time for pores with different number of atoms removed. The slope of the least-squared regression line of each curve is the water flux. (c) Two data augmentation techniques: flip and translation. (d) Water flux and ion rejection rate distribution of the final training dataset for predictive CNN model. Red dashed line represents a reverse sigmoid curve fitted in terms of water flux and ion rejection rate.}
    \label{fig:data}
\end{figure}

MD simulations were conducted using LAMMPS package\cite{plimpton1993fast}, where porous graphene membranes simulated were either created using Visual Molecular Dynamics\cite{humphrey1996vmd} or automatically generated by deep reinforcement agent (samples from early stage of training). A regular simulation system consists of four different sections: a graphene piston that applied constant external pressure; a saline water section containing potassium chloride as solute; a single-layer graphene membrane with the pore of different geometries; and a freshwater section which functioned as a reservoir of filtered water(Fig.~\ref{subfig:MD_system}). The molarity of the saline water in this work is  $\sim$2.28 M, which is higher than normal seawater for the sake of computational efficiency. The dimension of the simulation box is approximately 4 nm $\times$ 4 nm $\times$ 13 nm in ${x}$, ${y}$, and ${z}$ direction, respectively. A periodic boundary condition was applied to all three dimensions. 

All water molecules in this work were simulated using SPC/E model\cite{mark2001structure}, with SHAKE\cite{ryckaert1977numerical} algorithm to constrain the bond length and angles. Lennard-Jones (LJ) potentials (Supplementary Information Table S1) along with long-range Coulombic electrostatics potentials were adopted as interatomic potentials in the MD simulation. The cutoff for the interatomic potentials was set to be 12 $\si{\angstrom}$. Lorentz-Berthelot rules were employed for the calculation of LJ potentials between different kinds of atoms. Particle-particle particle-Mesh (PPPM) Ewald sovler\cite{plimpton1997particle} with 0.005 root-mean-squared error was used for long-range Coulombic potential correction. The porous graphene membrane and piston were each regarded as an entity during the simulation (internal interatomic potentials were not calculated) in order to reduce the computational cost.

In the first stage of each individual simulation, the internal energy of the system was minimized for 1000 iterations. The system then ran for 5 ps under the ${NPT}$ (isothermal-isobaric) ensemble at 300 K after the velocities of molecules were initialized based on Gaussian distribution. After the equilibrating, the system under ${NPT}$ ensemble, the system was switched to ${NVT}$ (canonical) ensemble to run for another 10 ns. The temperature was maintained at 300 K by Nos\'e-Hoover thermostat\cite{nose1984unified,hoover1985canonical} with a time constant of 0.5 ps. At this stage, a ${z}$-direction constant external pressure of 100 MPa was applied on saline water by the piston to mimic the reverse osmosis process in water desalination. Since the relationship between water flux and external pressure in reverse osmosis process was generally linear\cite{cohen2012water, heiranian2015water, cao2019water, cao2020single}, the performance of pores under 100 MPa could be extrapolated to lower pressures. Therefore, we chose to run simulations under 100 MPa external pressure to rapidly collect meaningful data. Molecular trajectories of each simulation were collected every 5 ps for data processing. 

\subsection{Dataset and Data Augmentation}
The two major performance indicators of a membrane in water desalination: water flux, and ion rejection rate, were calculated by post-processing the MD simulation trajectories. The slope of the fitted least square regression line on filtered water with respect to the simulation time curve was calculated to be the water flux of each membrane (Fig.~\ref{subfig:water_time}). The ion rejection rate of each membrane was calculated by dividing the number of ions in the freshwater section by the total number of ions.

The total number of different simulated porous graphene is 185. Since the reward of DRL agent in our model was calculated based on the water flux/ion rejection prediction of performance predictor (Eq.\ref{eqn:logistic} and \ref{eqn:reward}, Supplementary Information Fig.S2), highly accurate prediction must be achieved to ensure the quality of DRL training. A much larger training dataset was necessary for the optimization of CNN model. The method employed in our study to substantially increase the size of the dataset was data augmentation\cite{perez2017effectiveness, van2001art}. Given that the water desalination performance of a graphene pore depended on its size and geometry, we could assume that a flipped or translated pore on the same graphene membrane would demonstrate identical water flux/ion rejection rate of the original pore (Proven by MD simulations in Supplementary Information Fig.S3). Therefore, copies of original pores were created by being flipped along ${x}$ or ${y}$ axis and/or translating in -4$\si{\angstrom}$ to 4$\si{\angstrom}$ in ${x}$ and ${y}$ directions (Fig.~\ref{subfig:augmentation}). The water desalination performance of pore copies are random variable with normal distribution ($\mu$ = original pore performance, $\sigma$ = 1\% of original pore performance). In order to improve CNN's prediction accuracy on the performance of pores created by the DRL agent, we augmented DRL-generated pores 32 times. Among the other pores, the ones with zero water flux (too small to allow water transport) were augmented 6 times, and the rest of the pores were augmented 24 times. Data augmentation was conducted using the Atomic Simulation Environment (ASE) package\cite{larsen2017atomic}. The final dataset used for CNN training contains 3937 samples (Fig.~\ref{subfig:flux_rej}). A reverse sigmoid function was fitted to the distribution of samples to show the general relationship between the water flux and ion rejection rates.

\subsection{Water Desalination Performance Predictor}
To facilitate the reward estimation of DRL, a CNN model was trained to make an instantaneous prediction of water flux and ion rejection rates of a specific nanoporous graphene membrane. CNN is widely known as a universal feature extractor. Given that the water desalination performance of a graphene nanopore depends on its geometrical features, CNN can be the most suitable model to recognize geometrical features and make predictions based on them. There were 2 steps in the CNN modeling, including extracting features from the geometry of graphene nanoporous membrane and making predictions through a fully-connected multi-layer perceptron (MLP) regression model (Fig.\ref{fig:pipeline}). First of all, the geometrical features of a graphene nanoporous membrane is extracted to a 380 $\times$ 380 pixels representation. Color was applied on top of each atom, and all geometrical features were resized to the dimension of 224 $\times$ 224 pixels. The processed geometrical features was then fed into a CNN. Multiple CNN models, including ResNet18, ResNet50\cite{he2016deep}, and VGG16 \cite{simonyan2014very} with batch normalization were benchmarked based on the mean squared error (MSE) and ${R^2}$ of their resulting water flux/ion rejection rate predictions. An extracted feature vector with dimension of 1000 was output from the CNN model. Finally, given the feature vector, the MLP was able to make predictions of flux and ion rejection rates. The MLP used in this work consisted of 2 layers with 256 and 64 neurons in the first and second layer, respectively. A residual block\cite{he2016deep} and ReLU \cite{nair2010rectified} activation function were added after each layer of MLP. Two CNN models were trained: one for the prediction of water flux, and the other for ion rejection rate.

We compared the performance of CNN-based deep learning models with XGBoost \cite{chen2016xgboost}, a widely-used shallow machine learning model, was also trained to predict the water flux/ion rejection rate. The advantage of XGBoost model is that it requires much less of time for training compared to CNN. Before the training of XGBoost model, the graphene membrane was featurized into one-hot-encoded Morgan fingerprint \cite{rogers2010extended} vector of dimension 1024 using RDKit package\cite{landrum2016rdkit}, with a cutoff distance of 5$\si{\angstrom}$. The Morgan fingerprint vector was then fed in the XGBoost regression model as input. A random search was conducted on the hyperparameter grid (Supplementary Information Table S2) for model optimization.

In training the CNN models, we used gradient-based Adam\cite{kingma2014adam} optimizer with the learning rate $0.0001$ and $0.001$ for pretrained convolutional layers and the MLPs, respectively. To evaluate the performance of each model, the whole dataset was split into a training set and a test set with the ratio of 4:1. All models were trained only on the training set and tested on the test set. Raw values of water flux and ion rejection rate were standardized before fed into prediction models for training. CNN models were trained for 600 epochs, and the early stopping threshold of the XGBoost model was set as 1000 rounds. The model with the best performance was selected to be retrained on the whole dataset and utilized in the DRL framework.

\subsection{DRL Nanopore Optimization Agent}
Our goal was to design the optimal geometry of graphene nanopore for energy-efficient water desalination, which simultaneously demanded high flux and high ion rejection under certain external pressure. In order to optimize the nanopore, an agent was expected to remove atoms sequentially until a desired pore geometry was developed. To this end, the agent was set to interact with nanoporous graphene in a sequence of actions $a_t$, states $s_t$, and rewards $r_t$ within an episode of length $T$. The goal of the agent was to select the action such that it could maximize the future discounted return $R_t=\sum_{t=1}^T \gamma^{t-1} r_t$ in the finite Markov decision process (MDP) setting. In our case, we set the discount factor $\gamma$ to be 1. 

At timestep $t$, given the nanoporous graphene $G_t$, the agent observed the state $s_t$, which was composed of Morgan fingerprint \cite{rogers2010extended}, coordinates of all the atoms, along with CNN-extracted graphene geometrical features. The graphene geometry $g'_t$ was fed into the flux and ion rejection predictor, respectively. The geometrical features were the concatenation of last layer before output of the performance predictors. Once an atom was removed, its coordinate was set to the origin since MLP required a homogeneous input dimension. The predicted flux $f_t$ and ion rejection $i_t$ were leveraged to compute the reward signal $r_t$ for the agent, as given in Eq.~\ref{eqn:logistic} and Eq.~\ref{eqn:reward}:
\begin{linenomath}\begin{align}
    \sigma(x) &=  A + \frac{K-A}{(C + Q e^{-B x})^{\frac{1}{\nu}}},
    \label{eqn:logistic} \\
    r_t &= \alpha f_t + \sigma(i_t) - \sigma(1),
    \label{eqn:reward}
\end{align}\end{linenomath}
where $\sigma(\cdot)$ is the generalized logistic function\cite{richards1959flexible} and $\alpha$ is the coefficient for flux term. In our setting, $\alpha$ was set to be $0.01$, and $A=-15,\ K=0,\ B=13,\ Q=100,\ \nu=0.01,\ C=1$ for the logistic function. A linear term of flux reward encouraged the agent to expand nanopores, which would allow higher water flux. Since low ion rejection rate was not favored in water desalination, a generalized logistic function $\sigma(\cdot)$ was leveraged to penalize ion rejection term. When $i_t$ was high, $\sigma(i_t)$ was close to zero, allowing the growth of the nanopores. However, when $i_t$ was low, $\sigma(i_t)$ fiercely penalized the agent by outputting a large negative value(Supplementary Information Fig.S2). Besides, an extra $0.05$ reward was given to the agent when it chose to remove an atom at timestep $t$ to encourage pore growth at an early stage. Given state $s_t$ and reward $r_t$, the agent intended to choose the action $a_{t}$ for next step. However, due to the high dimensionality of possible action space (all the atoms in the graphene fragment), it was computationally expensive for the agent to efficiently and thoroughly explore the possible actions and to learn an optimal design. Therefore, only a subset of $M$ atoms was selected as candidates $c_{t}$. Atoms on the edge of pore were picked based on the rank of their proximity to the pore center, if the number exceeds $M$, only the first $M$ atoms closest to the center of pore were selected. However, when the number of edge atoms was less than $M$, non-edge atoms closest to the center of pore were selected as possible candidates to maintain the size of $c_{t}$. Given the state $s_t$, reward $r_t$, and candidate $c_t$, the agent learned to pick the action aiming to maximize future rewards.

To train the agent, deep Q-learning \cite{mnih2015human} with experience replay was implemented. Our task only considered deterministic environment, namely given $(s,c)$, the pair $(s',c')$ at the next time-step was unique. Based on Bellman equation \cite{sutton1998introduction}, the optimal action-value function $Q^*(s,c)$ in the deterministic environment was defined as:
\begin{linenomath}\begin{align}
    Q^*(s,c) = r + \gamma \max_{c'} Q^*(s',c')
\end{align}\end{linenomath}
To model the Q function, the Q-network parameterized by $\theta$ and target network parameterized by $\theta'$, two fully connected networks with the identical architecture were built. During training, only the parameters $\theta$ in the Q-network were updated through backpropagation from loss function. The parameters $\theta'$ in the target network were updated with $\theta$ every 10 steps and are kept fixed otherwise. The input to the network was the pair of graphene state and action candidates, $(s,c)$, and the output was the Q values of all the actions in the candidate. The agent then picked the action with the highest Q value. In addition, the agent's experience $(s, c, r, s')$ in the episodes were stored to a replay buffer $\mathcal{D}$ \cite{mnih2015human}, such that the experience can be leveraged to update the network parameters multiple times. During training, a mini-batch of samples was drawn uniformly at random from the replay buffer $(s, c, a, r, s') \sim U(\mathcal{D})$. The loss function (Eq.~\ref{eqn:loss}) measured the difference between the target Q value $Q^*(s',c';\theta_{i}')$ and the prediction of current Q network $Q(s,c;\theta_i)$:
\begin{linenomath}\begin{align}
    L_i(\theta_i) = \mathbb{E}_{(s,c,r,s')\sim U(\mathcal{D})} \bigg[ \Big( r+\gamma \max_{a'} Q(s',c'; \theta_{i}') - Q(s,c;\theta_i) \Big)^2 \bigg]
    \label{eqn:loss}
\end{align}\end{linenomath}
In our setting, we use Adam optimizer \cite{kingma2014adam} with learning rate $0.001$. The replay buffer is of capacity $10000$ and batch size is set to $128$.

\section{Results}
 The mean squared error (MSE) and coefficient of determination ($R{^2}$) are used as metrics to evaluate the performance predictions of models. The water flux and ion rejection labels are standardized before fed into the property prediction models. Thus the metrics tabulated are based on standardized water flux or ion rejection rate(Table~\ref{tb:predict}). Since the accuracy of performance predictor directly influence how accurately the DRL agent is rewarded/penalized during training, the model with least MSE and highest $R{^2}$ values was chosen to be used for reward estimation. ResNet \cite{he2016deep} significantly outperformed other models on both metrics, and the fined-tuned ResNet50 model reaches the highest accuracy in predicting both water flux and ion rejection rate. Therefore, a ResNet50 (retrained using the whole dataset) is used to predict the water desalination performance of various nanoporous graphenes to accelerate the DRL training. 

\begin{table}
  \centering
  \begin{tabular}{lllll}
    \hline
    Model & Flux MSE & Flux $R^2$ & Ion rejection MSE & Ion rejection $R^2$  \\
    \hline
    XGBoost \cite{chen2016xgboost}  & 0.011 & 0.988 & 0.008 & 0.992\\
    VGG16 \cite{simonyan2014very}   & 0.0448 & 0.957 & 0.0156 & 0.985\\
    ResNet18 \cite{he2016deep}      & 0.0024 & 0.998 & 0.0039 & 0.996 \\
    ResNet50 \cite{he2016deep}      & 0.0022 & 0.998 & 0.0038 & 0.996 \\
    \hline
  \end{tabular}
\caption{Performance of different models for graphene property prediction. Values based on standardized water flux/ion rejection.}
\label{tb:predict}
\end{table}

\begin{figure}[hp!]
    \begin{subfigure}{0.475\textwidth}
      \centering
      \includegraphics[width=0.9\linewidth]{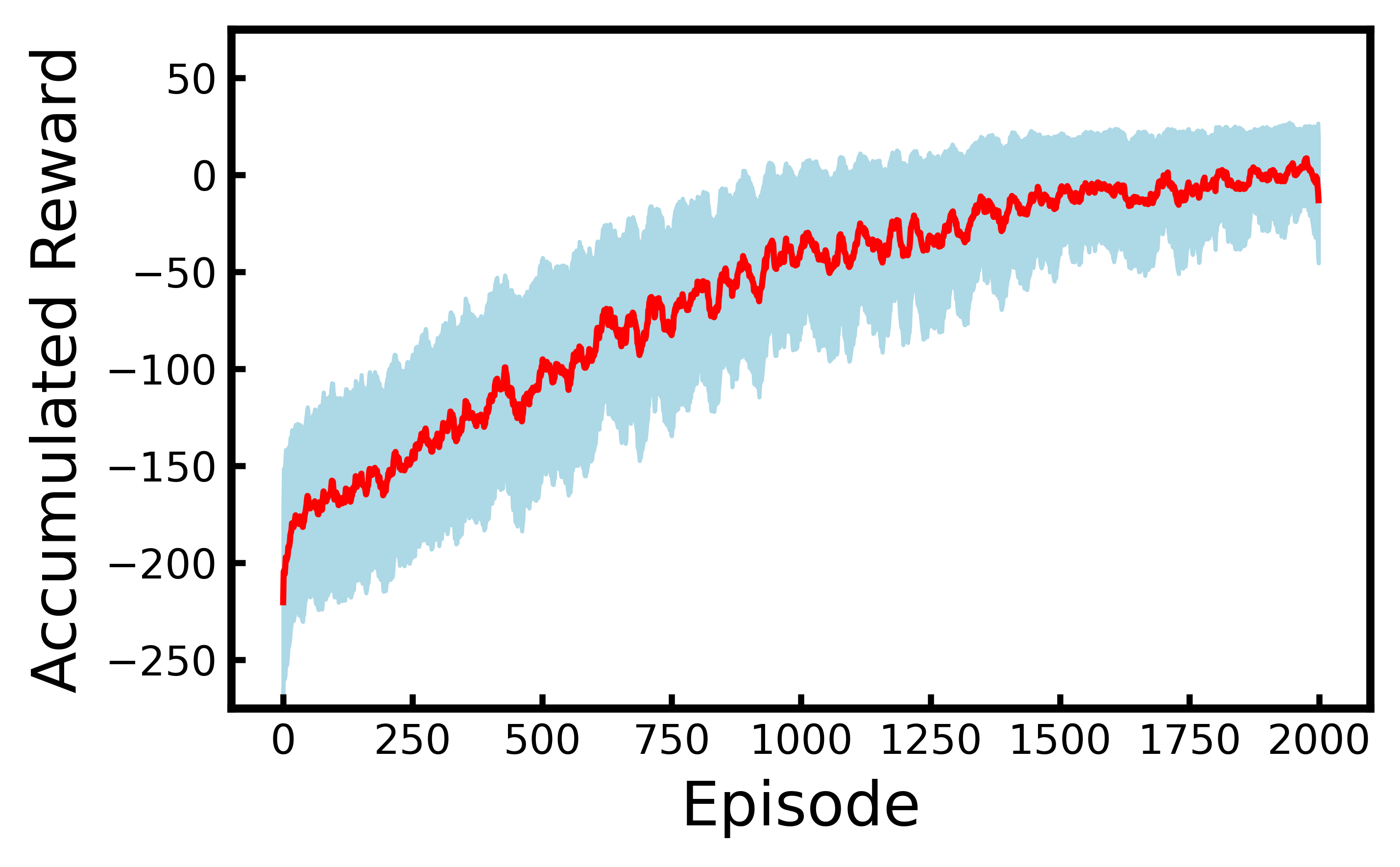}
      \caption{}
      \label{fig:rew_ep}
    \end{subfigure}
    \hfill
    \begin{subfigure}{0.475\textwidth}
      \centering
      \includegraphics[width=0.9\linewidth]{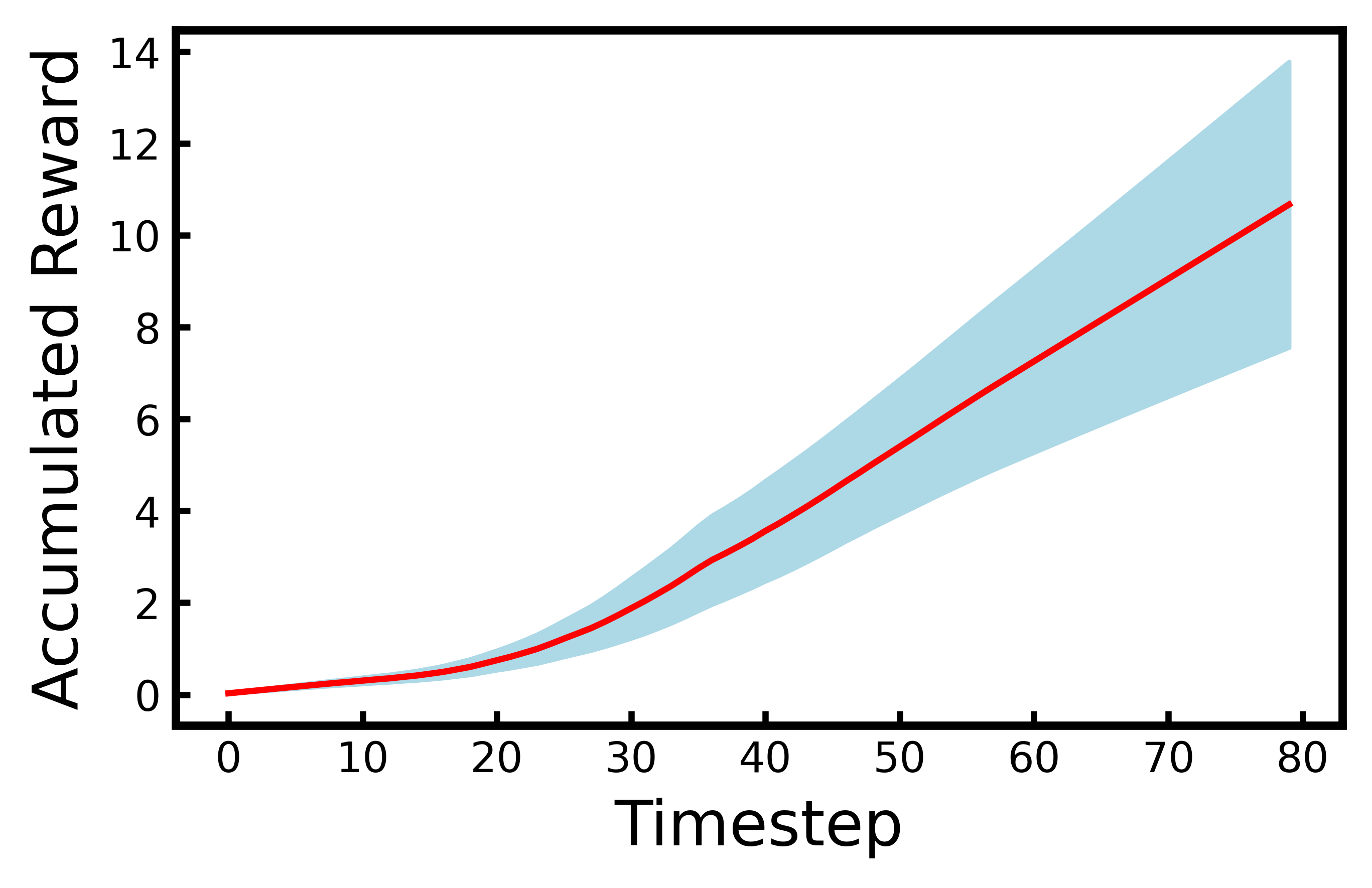}
      \caption{}
      \label{fig:rew_step}
    \end{subfigure}
    \vskip\baselineskip
    \begin{subfigure}{0.475\textwidth}
      \centering
      \includegraphics[width=0.9\linewidth]{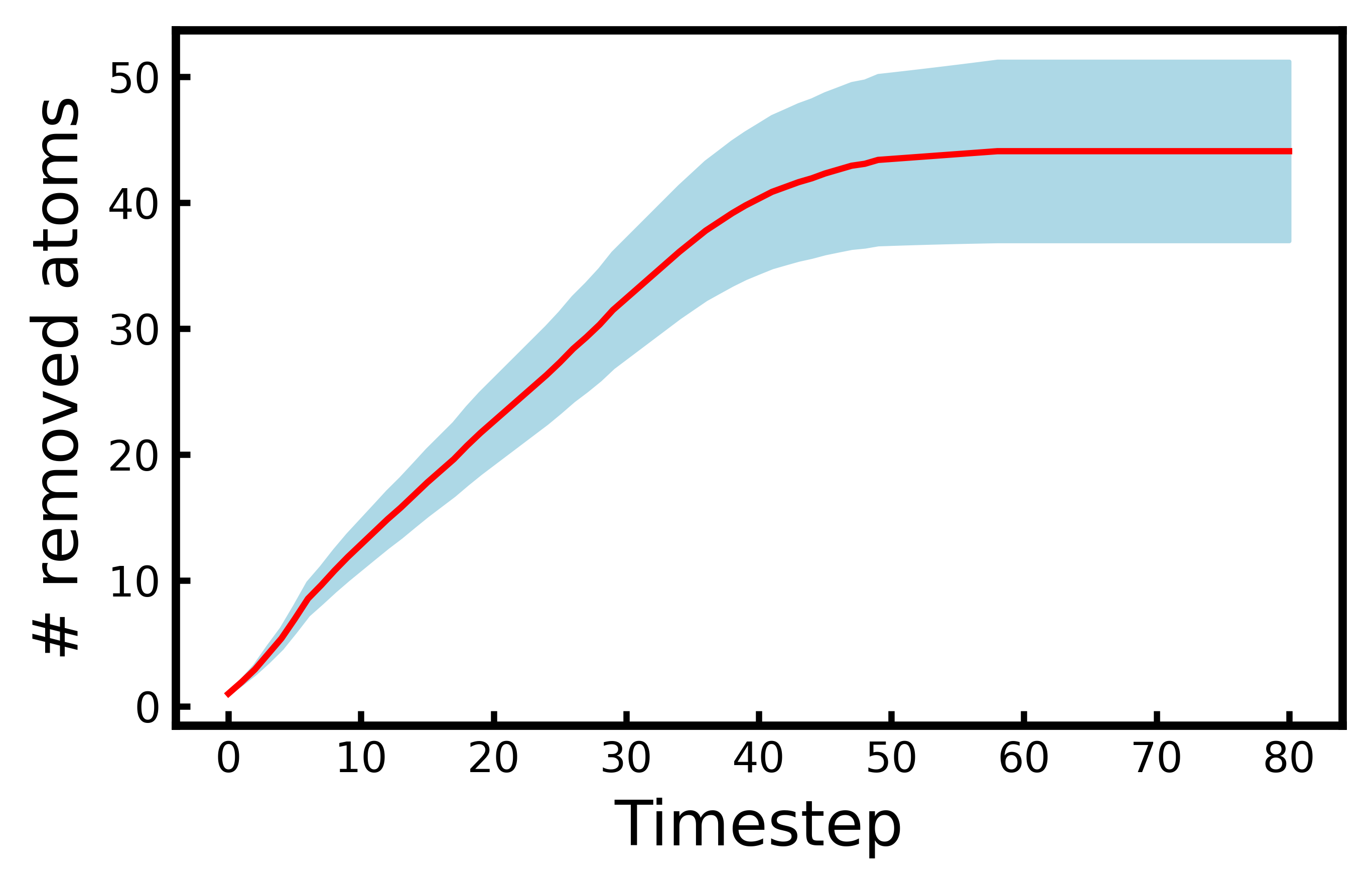}
      \caption{}
      \label{fig:num_rm_step}
    \end{subfigure}
    \hfill
    \begin{subfigure}{0.475\textwidth}
      \centering
      \includegraphics[width=0.9\linewidth]{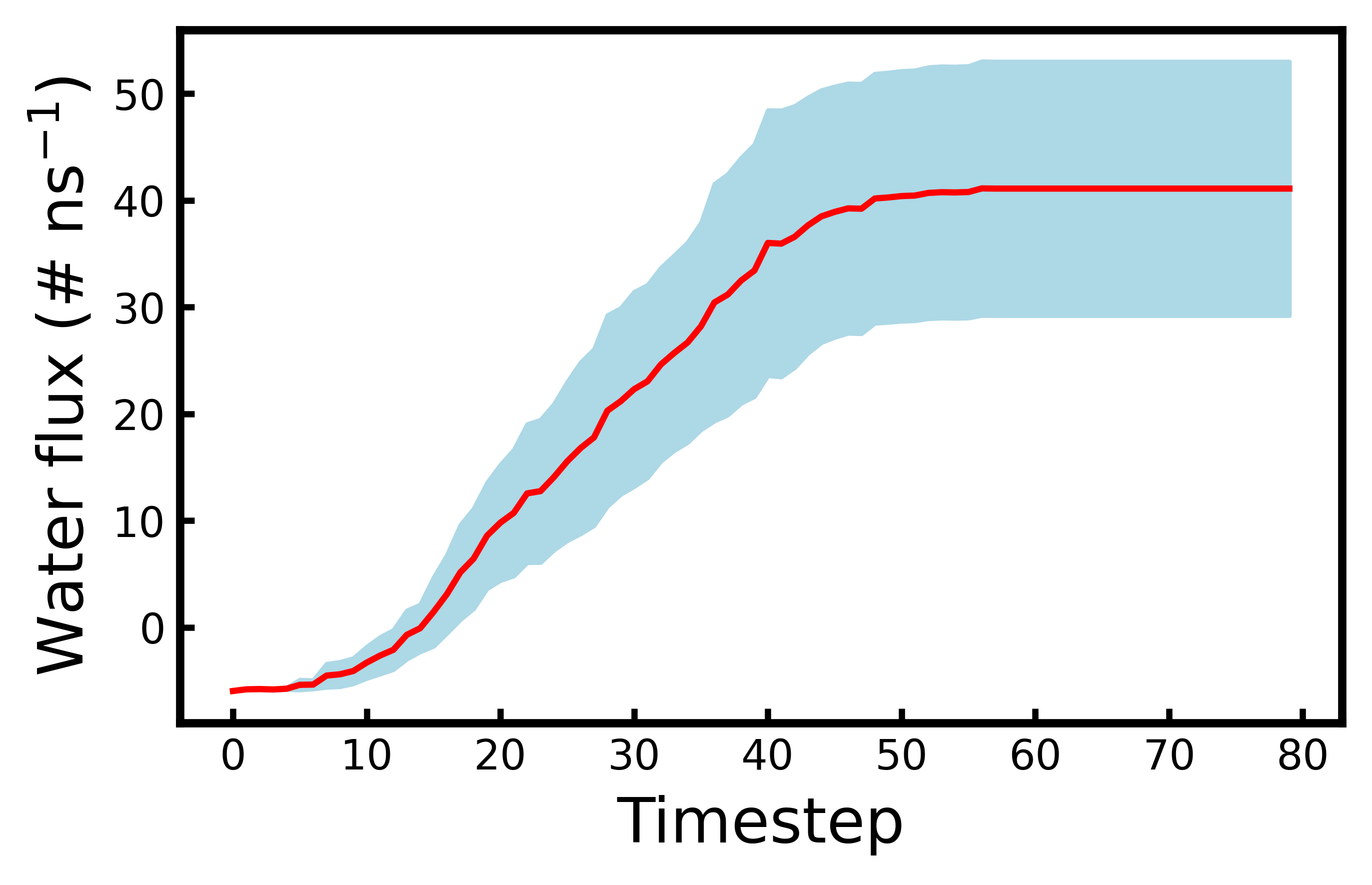}
      \caption{}
      \label{fig:flux_step}
    \end{subfigure}
    \vskip\baselineskip
    \begin{subfigure}{0.475\textwidth}
      \centering
      \includegraphics[width=0.9\linewidth]{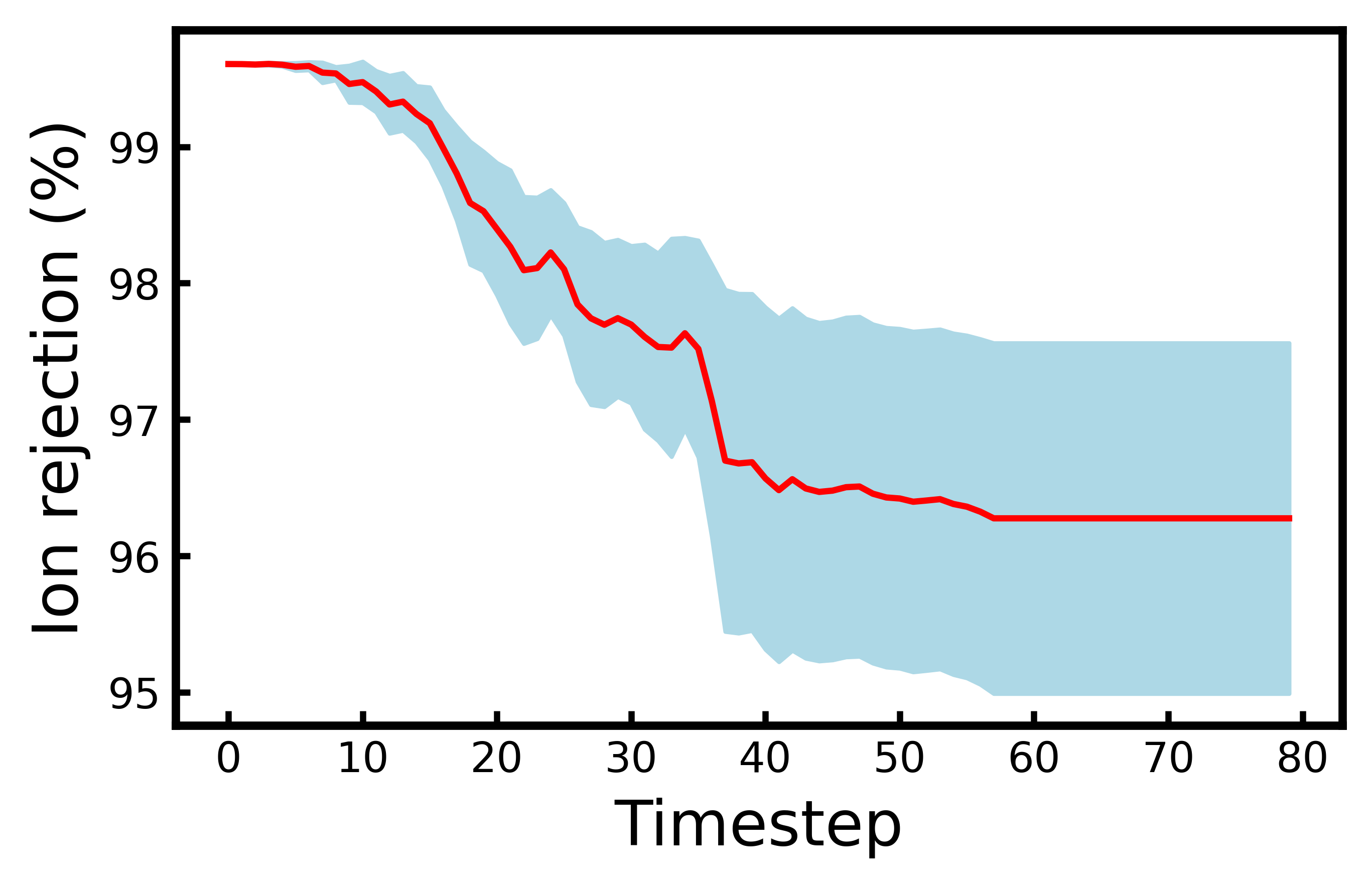}
      \caption{}
      \label{fig:rej_step}
    \end{subfigure}
    \hfill
    \begin{subfigure}{0.49\textwidth}
      \centering
      \includegraphics[width=0.7\linewidth]{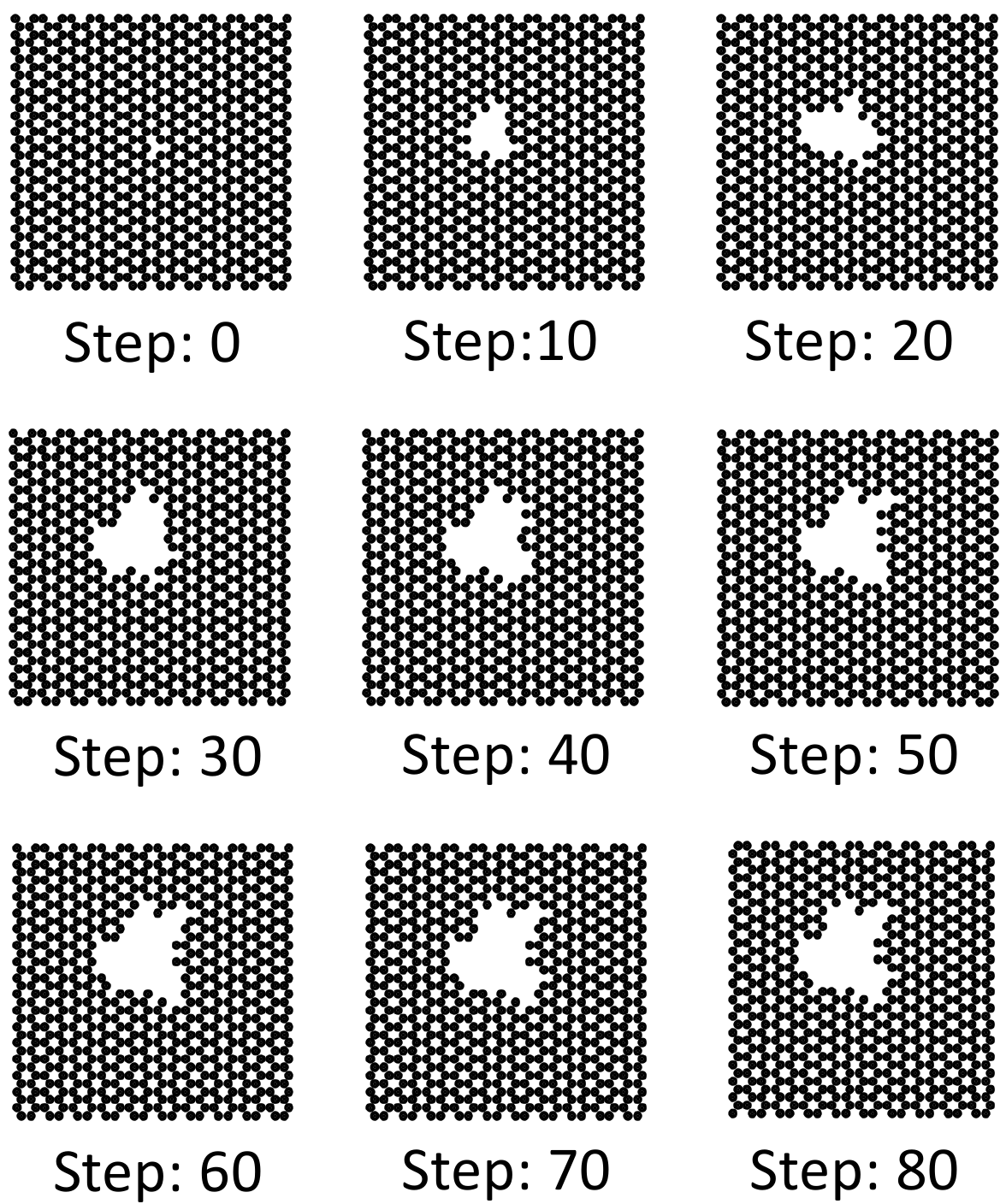}
      \caption{}
      \label{fig:evolve}
    \end{subfigure}
\caption{Training results for 10 DRL agents. (a) Summation of reward in each timestep vs. episode, where the red line is the running average of the reward with window size 21 and the blue shadow represents the standard deviation. (b) Summation of reward in each timestep vs. timestep. (c) Number of removed atoms vs. timestep. (d) Predicted water flux vs. timestep.  (e) Predicted ion rejection vs. timestep. Fig.~\ref{fig:rl_result}(b)-(e) show the results of DRL agents after trained for 2000 episodes, where the red line indicates the mean and the blue shadow is the standard deviation. (f) Evolution of a graphene nanopore designed by DRL agent.}
\label{fig:rl_result}
\end{figure}

We trained the DRL agent with 10 random seeds to generate various graphene nanopores. In the DRL agent training processes with different random seeds (Fig.\ref{fig:rl_result}), the red curves indicate mean values and the blue shadows represent standard deviations. The accumulated reward for each episode increases during training the DRL agent (Fig.~\ref{fig:rew_ep}). Initially the policy is noisy and the accumulated rewards are low, because the DRL agent has not yet learned to stop expanding the pore before receiving enormous penalty for low ion rejection rate. During the training, the DRL agent gradually learns a stable policy through maximizing the rewards (balancing the trade-off between water flux and ion rejection rate).The performance of DRL agent after 2000 episodes of training is demonstrated in Fig.~\ref{fig:rl_result}(b)-(e). The DRL agent generates nanopore which brings positive reward at each timestep, and the agent also automatically learns to stop enlarging the nanopore to avoid low ion rejection rate (Fig.~\ref{fig:rew_step} and ~\ref{fig:num_rm_step}). For example, the evolution of a DRL generated pore (Fig.~\ref{fig:evolve}, animated in Supplementary Movie) shows that DRL stops removing atom from the edge of graphene nanopore after 50th timestep, because it determines that higher water flux reward brought by further removing atoms is not worth the penalty for low ion rejection rate. Based on the prediction of the performance predictor, the DRL generated nanoporus graphenes have averaged $\sim$40 $\#/\si{\ns}$ water flux and $\sim$96\% ion rejection rate (Fig.~\ref{fig:flux_step} and ~\ref{fig:rej_step}).
\begin{figure}[]
    \begin{subfigure}{0.49\textwidth}
      \centering
      \includegraphics[width=\linewidth]{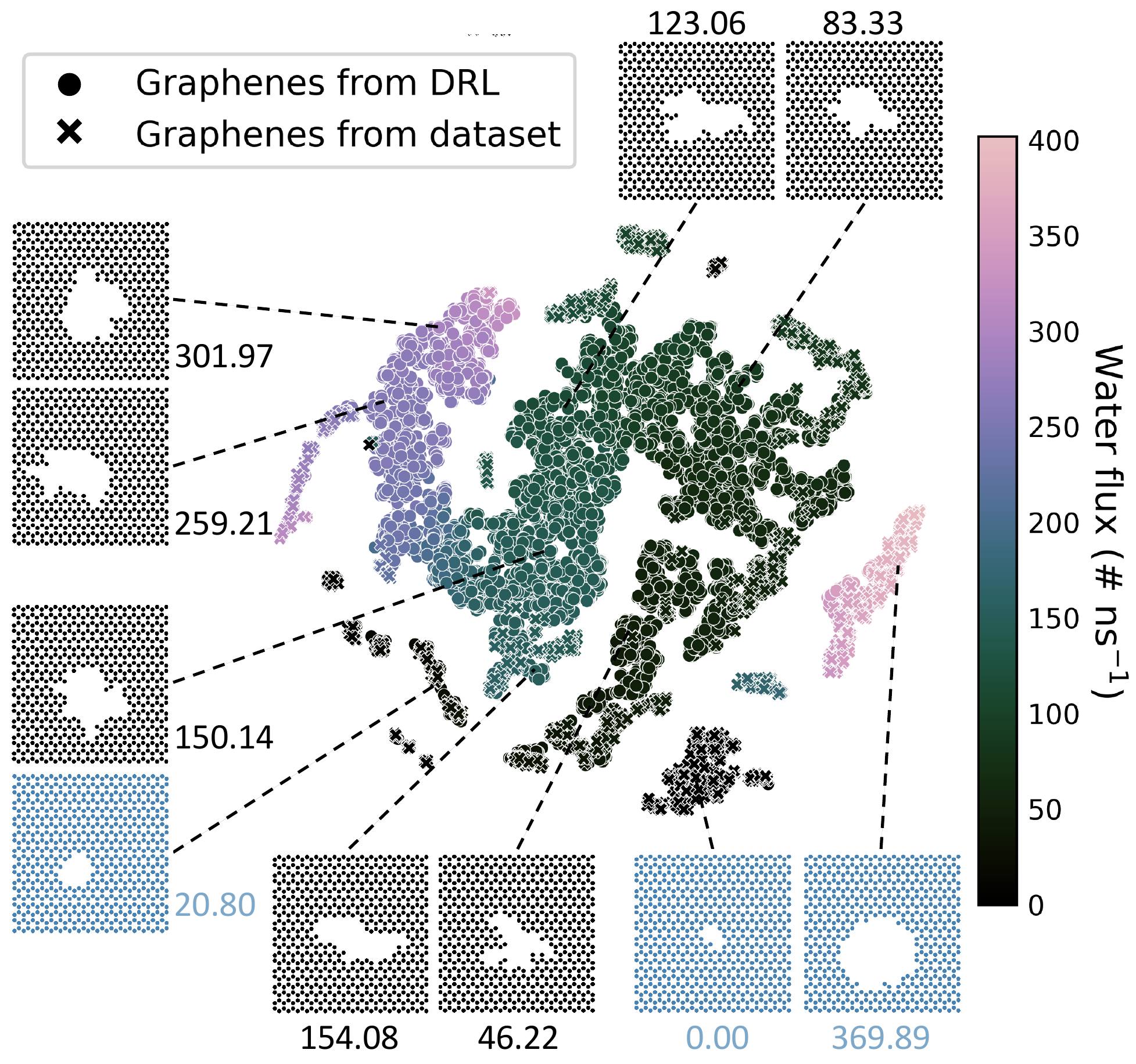}
      \caption{}
      \label{fig:tsne_flux}
    \end{subfigure}
    \hfill
    \begin{subfigure}{0.49\textwidth}
      \centering
      \includegraphics[width=\linewidth]{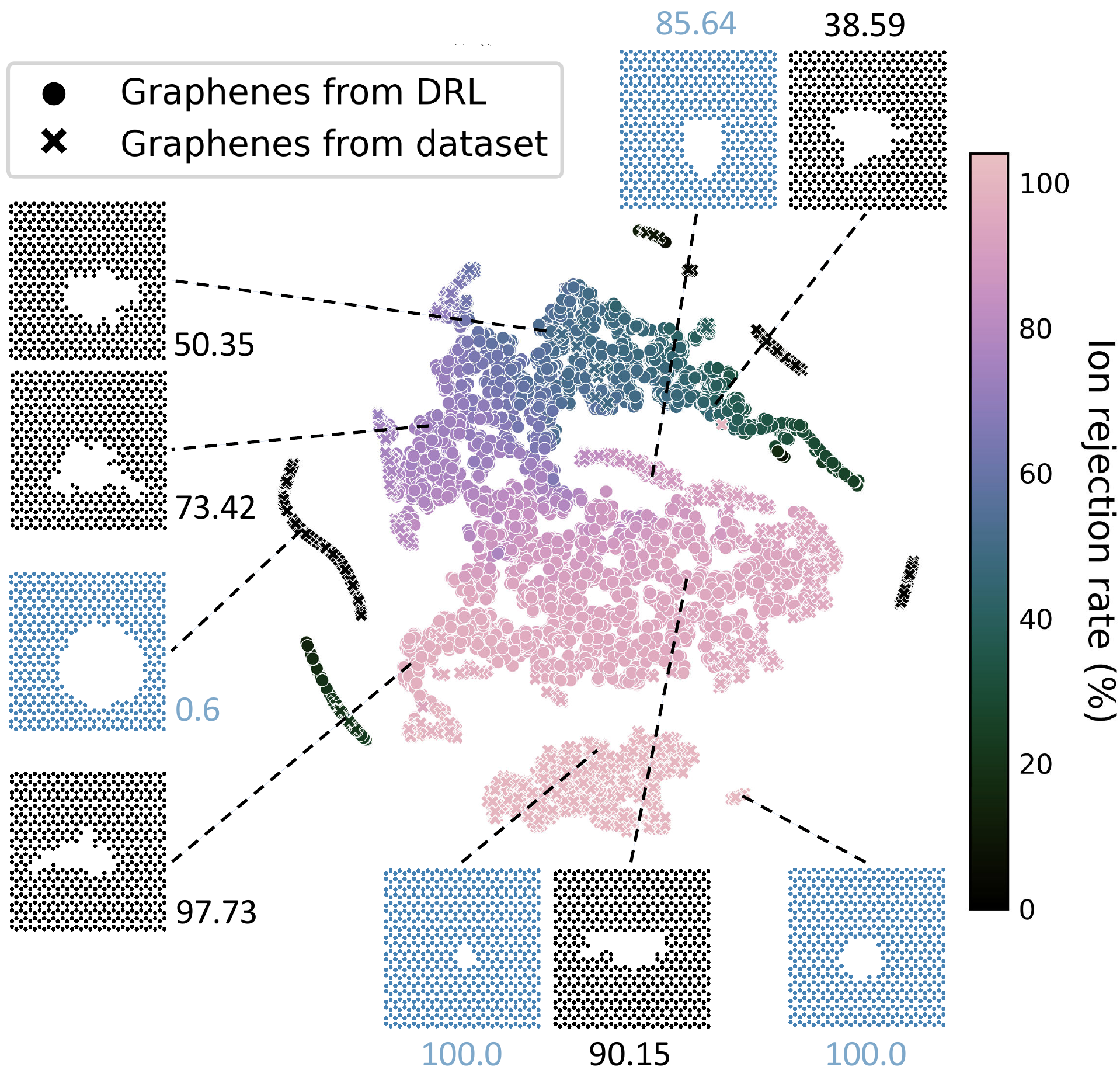}
      \caption{}
      \label{fig:tsne_rej}
    \end{subfigure}
\caption{(a) T-SNE\cite{maaten2008visualizing} visualization of features extracted from water flux prediction model. (b) T-SNE\cite{maaten2008visualizing} of features extracted from ion rejection rate prediction model. Black colored are DRL generated membranes and blue ones are from training dataset.}
\label{fig:flux_rej_rl_result}
\end{figure}

The collection of both DRL generated nanoporous graphene membranes (7999 samples) and membranes in the training dataset (3937 samples) is visualized using T-SNE\cite{maaten2008visualizing} algorithm (Fig.~\ref{fig:tsne_flux} and \ref{fig:tsne_rej}). T-SNE is a dimensionality reduction tool that is capable of mapping high-dimensional data to lower-dimension form while preserving the similarities between data points. In another words, membranes that are more similar to each other will have a higher tendency of being clustered. In this work, using CNN extracted features from each graphene membrane, T-SNE successfully clustered samples with similar water flux or ion rejection. This result indicates that features extracted from CNN models have a strong correlation with the water flux and ion rejection rate. 

\begin{figure}[]
  \begin{subfigure}{0.49\textwidth}
    \centering
    \includegraphics[width=0.9\linewidth]{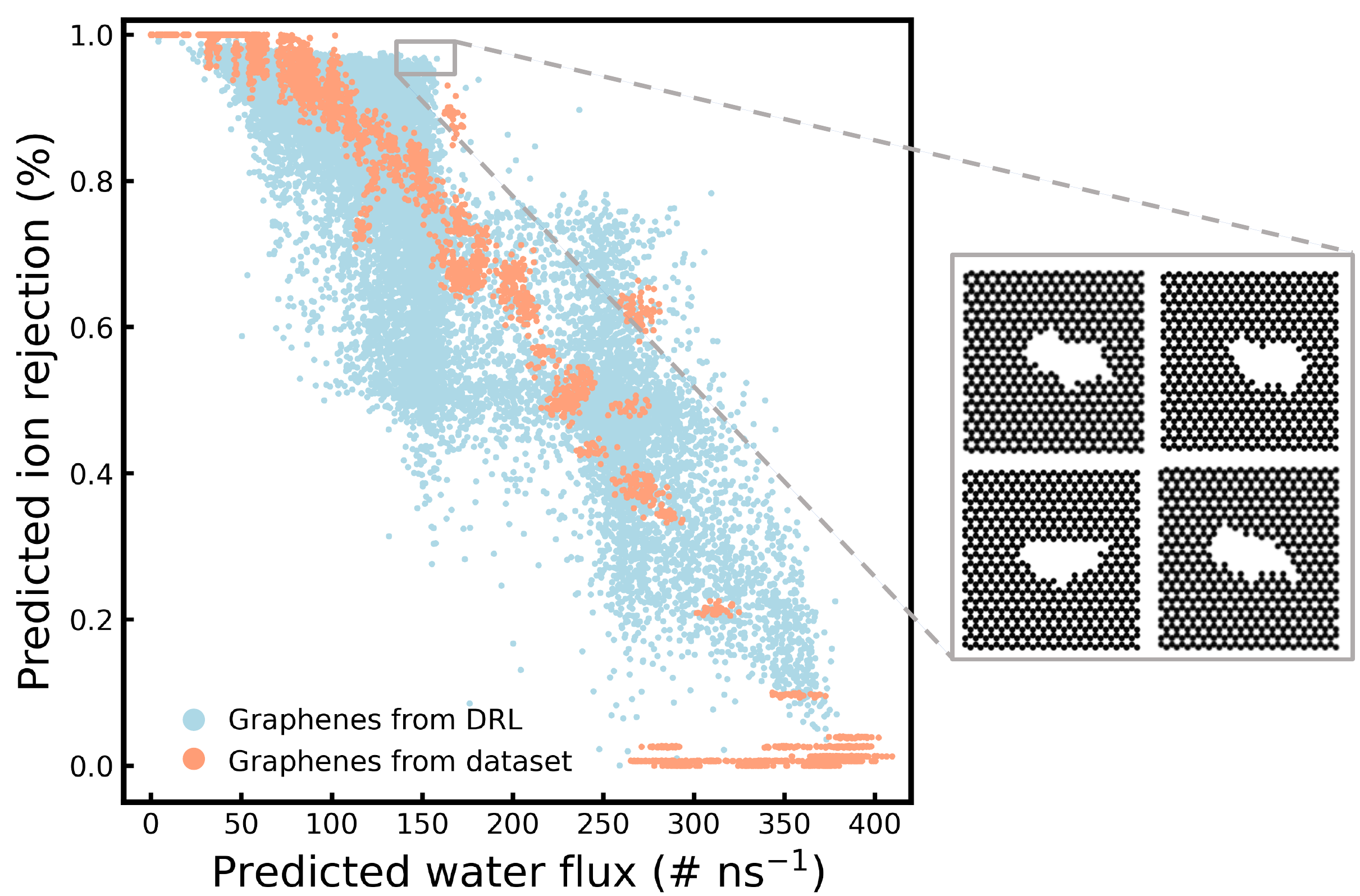}
    \caption{}
    \label{fig:flux_vs_rej}
  \end{subfigure}
  \hfill
  \begin{subfigure}{0.49\textwidth}
    \centering
    \includegraphics[width=0.9\linewidth]{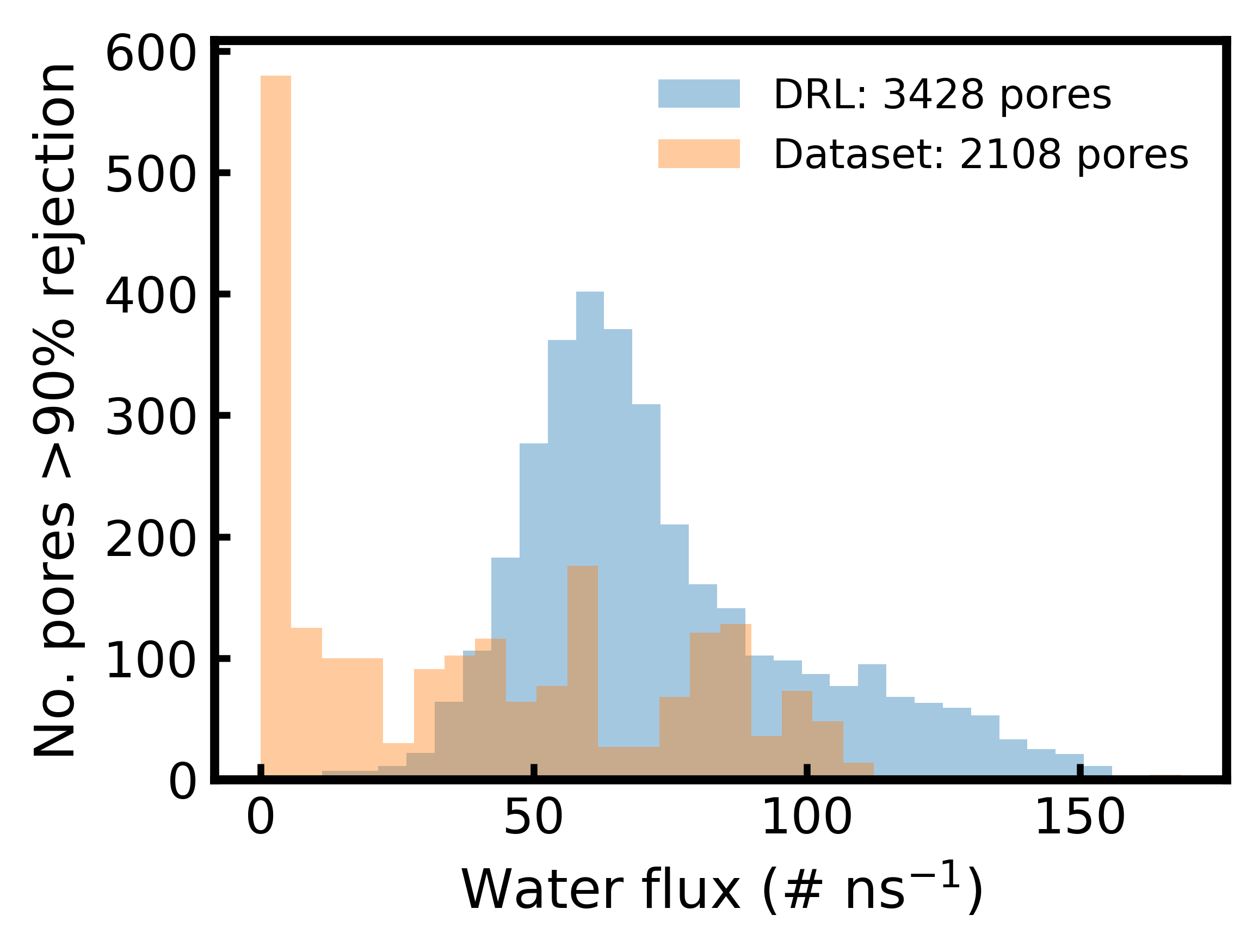}
    \caption{}
    \label{fig:flux_histo}
  \end{subfigure}

  \begin{subfigure}{0.45\textwidth}
    \centering
    \includegraphics[width=0.9\linewidth]{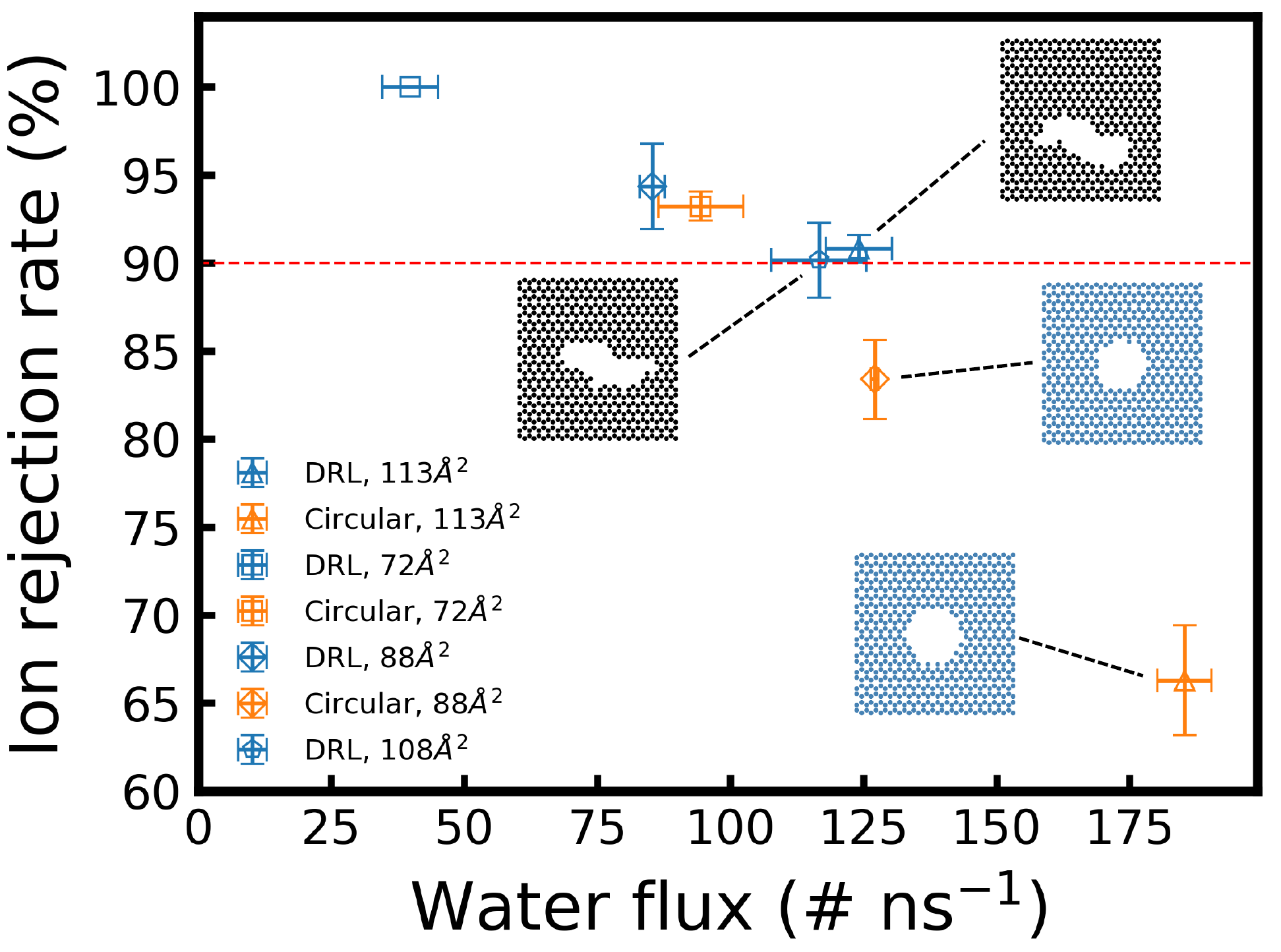}
    \caption{}
    \label{fig:DRL_vs_circular}
  \end{subfigure}
  \hfill
  \begin{subfigure}{0.45\textwidth}
    \centering
    \includegraphics[width=0.9\textwidth, keepaspectratio=true]{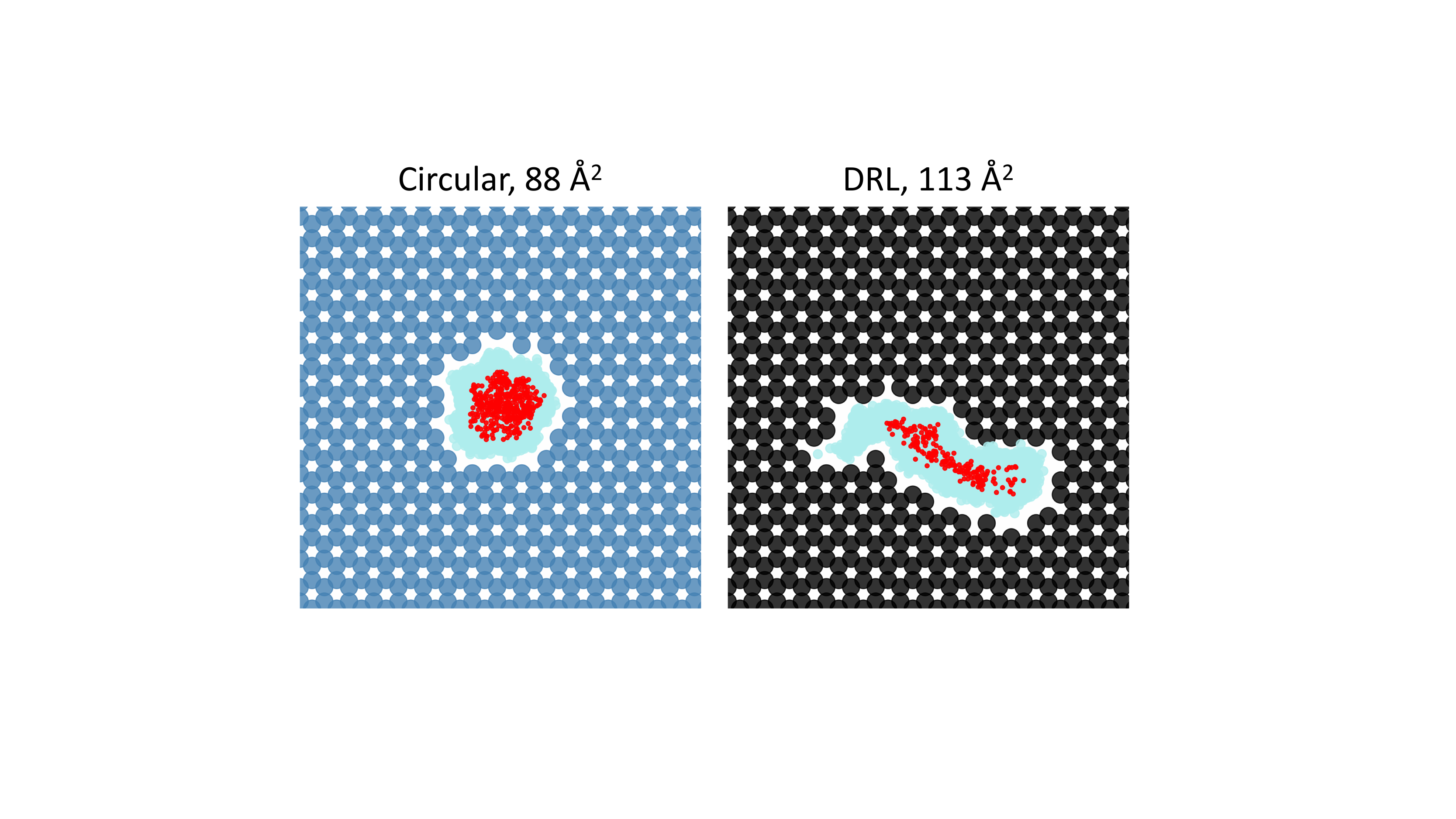}
    \caption{}
    \label{fig:ion_pass}
  \end{subfigure}
  
  \begin{subfigure}[b]{0.7\textwidth}
        \centering
        \includegraphics[width=0.9\textwidth, keepaspectratio=true]{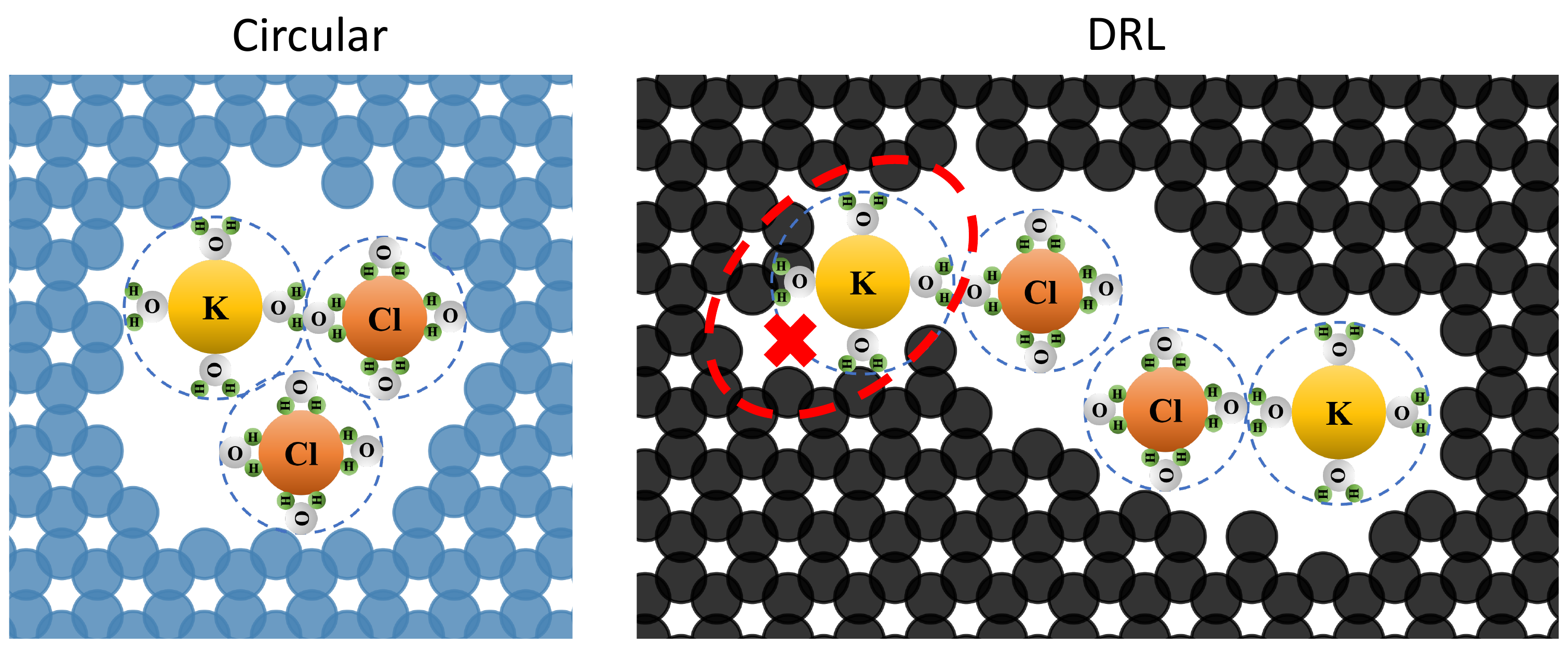}
        \caption{}
        \label{subfig:ion_free_zone}
    \end{subfigure}
  \caption{(a) Predicted water flux (ns$^{-1}$), and ion rejection rate (\%) of all graphene nanopores (7999 DRL generated + 3937 in training dataset). Zoom-in window shows the geometries of high-performance nanopores. (b) Histogram of water flux of nanopores with $>$90\% ion rejection rate. (c) Comparison of water desalination performance (under 100 MPa pressure) of circular and DRL generated graphene nanopores. Each data point is obtained by averaging the ion rejection and water flux of 4 MD simulations. The error bars represent one standard deviation. (d) Distribution of water molecules (aqua blue) and ions (red) when they are inside of the nanopores (Left: circular, area = 88$\si{\angstrom}^2$; Right: DRL generated, area = 113$\si{\angstrom}^2$). (e) Schematic showing how ions are blocked by DRL generated pore due to the steric effect. Ions with their hydration shell (Blue dashed circle) are too large to pass through the corner area (Red dashed circle) in the DRL generated pore, thus rendering that area an ion-free zone}
\label{fig:md_gen_graphene}
\end{figure}
The water desalination performances of all nanopores (DRL generated and training dataset) are compared in Fig.~\ref{fig:flux_vs_rej}. It is worth noting that the process of generating 7999 nanopores using DRL and predicting their water flux/ion rejection rate takes less than a single week; however, evaluating the performance of the same amount of nanopores using MD simulation will take $\sim$33 years (average 36 hours on each sample, using one 56-core CPU node). Among the nanopores with the same level of ion rejection rate, some nanopores discovered by DRL allow much higher water flux. One common feature shared by those high-performance nanopores is the semi-oval geometry with rough edges. We set 90\% ion rejection rate as the threshold to determine if a nanopore can effectively reject ions. The water flux histogram (Fig.~\ref{fig:flux_histo}) shows that given the baseline ion rejection rate as 90\%, DRL can extrapolate from the training dataset and discover graphene nanopores that generally allow higher water flux.

Further MD simulations are conducted with DRL generated membranes that show high predicted performances to evaluate how the DRL helps in optimizing graphene nanopore for water desalination (Simulation process recorded in Supplementary Movie). Although DRL generated pores generally have lower water flux compared with circular pores with the same area, they have much higher ion rejection rate (Fig.~\ref{fig:DRL_vs_circular}, 90\% threshold of ion rejection rate is marked by a red dashed line). For example, when the pore area is 113$\si{\angstrom}^2$, DRL generated nanopore maintained over 90\% ion rejection rate while the circular pore rejects only approximately 65\% of ions. A pore with high water flux but a very low ion rejection rate is not desirable in water desalination application. Moreover, the comparison between 113$\si{\angstrom}^2$ DRL generated nanopore with 88$\si{\angstrom}^2$ circular pore shows that DRL generated pore can reject more ions when achieving same water flux: they both have approximately 125 $\#/\si{\ns}$ water flux while 113$\si{\angstrom}^2$ DRL generated pore can reject approximately 7\% more ions. The comparison between simulation results proves that DRL tends to prioritize the ion rejection rate over water flux, which makes it capable of maximizing the water flux of nanopores while maintaining a valid ion rejection rate.

To gain deeper understanding about the reason behind the high ion rejection rate of DRL generated pores, distribution of water molecules and ions inside of 113$\si{\angstrom}^2$ DRL generated pore and 88$\si{\angstrom}^2$ circular pore have been visualized (Fig.~\ref{fig:ion_pass}). From the ion distribution (marked by red dots), we can observe that ions can traverse the circular pore evenly through the entire central area of the pore. The distributions of water molecules (marked by aqua blue color) and ions in the circular pore are in a homogeneous pattern. However, the corners inside of DRL generated nanopore are small enough to block the passage of ions while being large enough to accommodate the transport of water molecules. With the knowledge that ions are covered by hydration shell during the transport through nanopore, it can be seen that ion-free zones (corners) inside of DRL generated nanopore obstruct the traversing of ions with hydration shell by steric effect (Fig.~\ref{subfig:ion_free_zone}). This is the reason why high-performance nanopores (zoom-in Fig.~\ref{fig:flux_vs_rej}, more high-performance DRL-generated pores shown in Supplementary Information Fig.S5) all have rough edges. Discovers and utilizes this special geometry, DRL designs nanopores that can reject most ions while allowing high water transport.

\section{Conclusion}

In this work, we propose a graphene nanopore optimization framework based on the DRL agent accelerated by the water desalination performance predictor. In particular, we focus on the optimal design of nanoporous graphene for water desalination. The DRL agent takes the current graphene geometrical features and the candidate atoms as inputs to determine which atom to remove at each timestep. Trained with the DQN algorithm, the agent learns to generate nanopores that allow high water flux while maintaining high ion rejection. ResNet50, a widely used CNN model, is trained on a nanoporous graphene dataset to instantly predict the water flux and ion rejection rate under certain pressure. Such prediction by the ResNet50 enables the real-time interaction between the DRL agent with the graphene nanopores, as well as the online optimization of the DRL agent. CNN accelerated DRL training significantly expedites the exploration of graphene nanopores: 7999 different nanopores are designed and evaluated for water desalination performance during one-week training of DRL. Evaluating the same amount of graphene nanopores using MD simulation can take approximately 33 years with a 56-cores CPU cluster. When we set the baseline ion rejection rate to be 90\%, DRL shows the capability of extrapolating from existing training dataset to discover nanopore with higher water flux. Further MD simulations confirm that DRL generated nanopores outperform circular nanopores in term of ion rejection rate when they have approximately the same water flux. The better water desalination performance of DRL generated pores can be attributed to DRL's utilization of rough edges and small corners to block ions with hydration shell. In conclusion, DRL shows the capability of designing optimal graphene nanopores for water desalination. Moreover, with only minor modification, this framework can be directly extended to many other fields concerning nanopore design. With a well-trained machine learning property predictor, the DRL can automatically learn to design the optimal material structure effectively and efficiently.

\begin{acknowledgement}
The authors thank the supercomputing resource Arjuna provided by the Pittsburgh Supercomputing Center (PSC). This work is supported by
the start-up fund provided by CMU Mechanical Engineering.
\end{acknowledgement}

% \section{Author Contributions}
% A.B.F conceived the project, Y.W and Z.C designed and coded the DRL architecture, Y.W coded the CNN performance predictor, Z.C performed the MD simulations, all authors wrote the manuscript.

% \section{Competing Interests}
% The authors declare no competing financial or non-financial interests.

% \section{Data Availability}
% The data that support the findings of this study are available from the corresponding author upon reasonable request.
% \newpage

\newpage
\nocite{*}
\mciteErrorOnUnknownfalse
\bibliography{reference}

\providecommand{\latin}[1]{#1}
\makeatletter
\providecommand{\doi}
  {\begingroup\let\do\@makeother\dospecials
  \catcode`\{=1 \catcode`\}=2 \doi@aux}
\providecommand{\doi@aux}[1]{\endgroup\texttt{#1}}
\makeatother
\providecommand*\mcitethebibliography{\thebibliography}
\csname @ifundefined\endcsname{endmcitethebibliography}
  {\let\endmcitethebibliography\endthebibliography}{}
\begin{mcitethebibliography}{51}
\providecommand*\natexlab[1]{#1}
\providecommand*\mciteSetBstSublistMode[1]{}
\providecommand*\mciteSetBstMaxWidthForm[2]{}
\providecommand*\mciteBstWouldAddEndPuncttrue
  {\def\EndOfBibitem{\unskip.}}
\providecommand*\mciteBstWouldAddEndPunctfalse
  {\let\EndOfBibitem\relax}
\providecommand*\mciteSetBstMidEndSepPunct[3]{}
\providecommand*\mciteSetBstSublistLabelBeginEnd[3]{}
\providecommand*\EndOfBibitem{}
\mciteSetBstSublistMode{f}
\mciteSetBstMaxWidthForm{subitem}{(\alph{mcitesubitemcount})}
\mciteSetBstSublistLabelBeginEnd
  {\mcitemaxwidthsubitemform\space}
  {\relax}
  {\relax}

\bibitem[Jiang \latin{et~al.}(2009)Jiang, Cooper, and Dai]{jiang2009porous}
Jiang,~D.-e.; Cooper,~V.~R.; Dai,~S. Porous graphene as the ultimate membrane
  for gas separation. \emph{Nano letters} \textbf{2009}, \emph{9},
  4019--4024\relax
\mciteBstWouldAddEndPuncttrue
\mciteSetBstMidEndSepPunct{\mcitedefaultmidpunct}
{\mcitedefaultendpunct}{\mcitedefaultseppunct}\relax
\EndOfBibitem
\bibitem[Li \latin{et~al.}(2013)Li, Song, Zhang, Huang, Li, Mao, Ploehn, Bao,
  and Yu]{li2013ultrathin}
Li,~H.; Song,~Z.; Zhang,~X.; Huang,~Y.; Li,~S.; Mao,~Y.; Ploehn,~H.~J.;
  Bao,~Y.; Yu,~M. Ultrathin, molecular-sieving graphene oxide membranes for
  selective hydrogen separation. \emph{Science} \textbf{2013}, \emph{342},
  95--98\relax
\mciteBstWouldAddEndPuncttrue
\mciteSetBstMidEndSepPunct{\mcitedefaultmidpunct}
{\mcitedefaultendpunct}{\mcitedefaultseppunct}\relax
\EndOfBibitem
\bibitem[Kim \latin{et~al.}(2013)Kim, Yoon, Yoon, Yoo, Ahn, Cho, Shin, Yang,
  Paik, Kwon, \latin{et~al.} others]{kim2013selective}
Kim,~H.~W.; Yoon,~H.~W.; Yoon,~S.-M.; Yoo,~B.~M.; Ahn,~B.~K.; Cho,~Y.~H.;
  Shin,~H.~J.; Yang,~H.; Paik,~U.; Kwon,~S., \latin{et~al.}  Selective gas
  transport through few-layered graphene and graphene oxide membranes.
  \emph{Science} \textbf{2013}, \emph{342}, 91--95\relax
\mciteBstWouldAddEndPuncttrue
\mciteSetBstMidEndSepPunct{\mcitedefaultmidpunct}
{\mcitedefaultendpunct}{\mcitedefaultseppunct}\relax
\EndOfBibitem
\bibitem[Wang \latin{et~al.}(2009)Wang, Shi, Huang, Ma, Wang, Chen, and
  Chen]{wang2009supercapacitor}
Wang,~Y.; Shi,~Z.; Huang,~Y.; Ma,~Y.; Wang,~C.; Chen,~M.; Chen,~Y.
  Supercapacitor devices based on graphene materials. \emph{The Journal of
  Physical Chemistry C} \textbf{2009}, \emph{113}, 13103--13107\relax
\mciteBstWouldAddEndPuncttrue
\mciteSetBstMidEndSepPunct{\mcitedefaultmidpunct}
{\mcitedefaultendpunct}{\mcitedefaultseppunct}\relax
\EndOfBibitem
\bibitem[Liu \latin{et~al.}(2010)Liu, Yu, Neff, Zhamu, and
  Jang]{liu2010graphene}
Liu,~C.; Yu,~Z.; Neff,~D.; Zhamu,~A.; Jang,~B.~Z. Graphene-based supercapacitor
  with an ultrahigh energy density. \emph{Nano letters} \textbf{2010},
  \emph{10}, 4863--4868\relax
\mciteBstWouldAddEndPuncttrue
\mciteSetBstMidEndSepPunct{\mcitedefaultmidpunct}
{\mcitedefaultendpunct}{\mcitedefaultseppunct}\relax
\EndOfBibitem
\bibitem[Farimani \latin{et~al.}(2014)Farimani, Min, and
  Aluru]{farimani2014dna}
Farimani,~A.~B.; Min,~K.; Aluru,~N.~R. DNA base detection using a single-layer
  MoS2. \emph{ACS nano} \textbf{2014}, \emph{8}, 7914--7922\relax
\mciteBstWouldAddEndPuncttrue
\mciteSetBstMidEndSepPunct{\mcitedefaultmidpunct}
{\mcitedefaultendpunct}{\mcitedefaultseppunct}\relax
\EndOfBibitem
\bibitem[Barati~Farimani \latin{et~al.}(2017)Barati~Farimani, Dibaeinia, and
  Aluru]{barati2017dna}
Barati~Farimani,~A.; Dibaeinia,~P.; Aluru,~N.~R. DNA Origami--Graphene Hybrid
  Nanopore for DNA Detection. \emph{ACS Applied Materials \& Interfaces}
  \textbf{2017}, \emph{9}, 92--100\relax
\mciteBstWouldAddEndPuncttrue
\mciteSetBstMidEndSepPunct{\mcitedefaultmidpunct}
{\mcitedefaultendpunct}{\mcitedefaultseppunct}\relax
\EndOfBibitem
\bibitem[Schneider \latin{et~al.}(2010)Schneider, Kowalczyk, Calado, Pandraud,
  Zandbergen, Vandersypen, and Dekker]{schneider2010dna}
Schneider,~G.~F.; Kowalczyk,~S.~W.; Calado,~V.~E.; Pandraud,~G.;
  Zandbergen,~H.~W.; Vandersypen,~L.~M.; Dekker,~C. DNA translocation through
  graphene nanopores. \emph{Nano letters} \textbf{2010}, \emph{10},
  3163--3167\relax
\mciteBstWouldAddEndPuncttrue
\mciteSetBstMidEndSepPunct{\mcitedefaultmidpunct}
{\mcitedefaultendpunct}{\mcitedefaultseppunct}\relax
\EndOfBibitem
\bibitem[Cohen-Tanugi and Grossman(2012)Cohen-Tanugi, and
  Grossman]{cohen2012water}
Cohen-Tanugi,~D.; Grossman,~J.~C. Water desalination across nanoporous
  graphene. \emph{Nano letters} \textbf{2012}, \emph{12}, 3602--3608\relax
\mciteBstWouldAddEndPuncttrue
\mciteSetBstMidEndSepPunct{\mcitedefaultmidpunct}
{\mcitedefaultendpunct}{\mcitedefaultseppunct}\relax
\EndOfBibitem
\bibitem[Surwade \latin{et~al.}(2015)Surwade, Smirnov, Vlassiouk, Unocic,
  Veith, Dai, and Mahurin]{surwade2015water}
Surwade,~S.~P.; Smirnov,~S.~N.; Vlassiouk,~I.~V.; Unocic,~R.~R.; Veith,~G.~M.;
  Dai,~S.; Mahurin,~S.~M. Water desalination using nanoporous single-layer
  graphene. \emph{Nature nanotechnology} \textbf{2015}, \emph{10},
  459--464\relax
\mciteBstWouldAddEndPuncttrue
\mciteSetBstMidEndSepPunct{\mcitedefaultmidpunct}
{\mcitedefaultendpunct}{\mcitedefaultseppunct}\relax
\EndOfBibitem
\bibitem[Heiranian \latin{et~al.}(2015)Heiranian, Farimani, and
  Aluru]{heiranian2015water}
Heiranian,~M.; Farimani,~A.~B.; Aluru,~N.~R. Water desalination with a
  single-layer MoS 2 nanopore. \emph{Nature communications} \textbf{2015},
  \emph{6}, 1--6\relax
\mciteBstWouldAddEndPuncttrue
\mciteSetBstMidEndSepPunct{\mcitedefaultmidpunct}
{\mcitedefaultendpunct}{\mcitedefaultseppunct}\relax
\EndOfBibitem
\bibitem[Cao \latin{et~al.}(2019)Cao, Liu, and Barati~Farimani]{cao2019water}
Cao,~Z.; Liu,~V.; Barati~Farimani,~A. Water Desalination with Two-Dimensional
  Metal--Organic Framework Membranes. \emph{Nano letters} \textbf{2019},
  \emph{19}, 8638--8643\relax
\mciteBstWouldAddEndPuncttrue
\mciteSetBstMidEndSepPunct{\mcitedefaultmidpunct}
{\mcitedefaultendpunct}{\mcitedefaultseppunct}\relax
\EndOfBibitem
\bibitem[Cao \latin{et~al.}(2020)Cao, Liu, and Barati~Farimani]{cao2020single}
Cao,~Z.; Liu,~V.; Barati~Farimani,~A. Why Single-layer MoS2 is a More Energy
  Efficient Membrane for Water Desalination? \emph{ACS Energy Letters}
  \textbf{2020}, \relax
\mciteBstWouldAddEndPunctfalse
\mciteSetBstMidEndSepPunct{\mcitedefaultmidpunct}
{}{\mcitedefaultseppunct}\relax
\EndOfBibitem
\bibitem[LeCun \latin{et~al.}(2015)LeCun, Bengio, and Hinton]{lecun2015deep}
LeCun,~Y.; Bengio,~Y.; Hinton,~G. Deep learning. \emph{nature} \textbf{2015},
  \emph{521}, 436--444\relax
\mciteBstWouldAddEndPuncttrue
\mciteSetBstMidEndSepPunct{\mcitedefaultmidpunct}
{\mcitedefaultendpunct}{\mcitedefaultseppunct}\relax
\EndOfBibitem
\bibitem[Mnih \latin{et~al.}(2015)Mnih, Kavukcuoglu, Silver, Rusu, Veness,
  Bellemare, Graves, Riedmiller, Fidjeland, Ostrovski, \latin{et~al.}
  others]{mnih2015human}
Mnih,~V.; Kavukcuoglu,~K.; Silver,~D.; Rusu,~A.~A.; Veness,~J.;
  Bellemare,~M.~G.; Graves,~A.; Riedmiller,~M.; Fidjeland,~A.~K.;
  Ostrovski,~G., \latin{et~al.}  Human-level control through deep reinforcement
  learning. \emph{Nature} \textbf{2015}, \emph{518}, 529--533\relax
\mciteBstWouldAddEndPuncttrue
\mciteSetBstMidEndSepPunct{\mcitedefaultmidpunct}
{\mcitedefaultendpunct}{\mcitedefaultseppunct}\relax
\EndOfBibitem
\bibitem[Sutton \latin{et~al.}(1998)Sutton, Barto, \latin{et~al.}
  others]{sutton1998introduction}
Sutton,~R.~S.; Barto,~A.~G., \latin{et~al.}  \emph{Introduction to
  reinforcement learning}; MIT press Cambridge, 1998; Vol. 135\relax
\mciteBstWouldAddEndPuncttrue
\mciteSetBstMidEndSepPunct{\mcitedefaultmidpunct}
{\mcitedefaultendpunct}{\mcitedefaultseppunct}\relax
\EndOfBibitem
\bibitem[Mnih \latin{et~al.}(2013)Mnih, Kavukcuoglu, Silver, Graves,
  Antonoglou, Wierstra, and Riedmiller]{mnih2013playing}
Mnih,~V.; Kavukcuoglu,~K.; Silver,~D.; Graves,~A.; Antonoglou,~I.;
  Wierstra,~D.; Riedmiller,~M. Playing atari with deep reinforcement learning.
  \emph{arXiv preprint arXiv:1312.5602} \textbf{2013}, \relax
\mciteBstWouldAddEndPunctfalse
\mciteSetBstMidEndSepPunct{\mcitedefaultmidpunct}
{}{\mcitedefaultseppunct}\relax
\EndOfBibitem
\bibitem[Popova \latin{et~al.}(2018)Popova, Isayev, and
  Tropsha]{popova2018deep}
Popova,~M.; Isayev,~O.; Tropsha,~A. Deep reinforcement learning for de novo
  drug design. \emph{Science advances} \textbf{2018}, \emph{4}, eaap7885\relax
\mciteBstWouldAddEndPuncttrue
\mciteSetBstMidEndSepPunct{\mcitedefaultmidpunct}
{\mcitedefaultendpunct}{\mcitedefaultseppunct}\relax
\EndOfBibitem
\bibitem[Karamad \latin{et~al.}(2020)Karamad, Magar, Shi, Siahrostami, Gates,
  and Farimani]{karamad2020orbital}
Karamad,~M.; Magar,~R.; Shi,~Y.; Siahrostami,~S.; Gates,~I.~D.; Farimani,~A.~B.
  Orbital Graph Convolutional Neural Network for Material Property Prediction.
  \emph{arXiv preprint arXiv:2008.06415} \textbf{2020}, \relax
\mciteBstWouldAddEndPunctfalse
\mciteSetBstMidEndSepPunct{\mcitedefaultmidpunct}
{}{\mcitedefaultseppunct}\relax
\EndOfBibitem
\bibitem[Yao \latin{et~al.}(2020)Yao, Sanchez-Lengeling, Bobbitt, Bucior,
  Kumar, Collins, Burns, Woo, Farha, Snurr, \latin{et~al.}
  others]{yao2020inverse}
Yao,~Z.; Sanchez-Lengeling,~B.; Bobbitt,~N.~S.; Bucior,~B.~J.; Kumar,~S. G.~H.;
  Collins,~S.~P.; Burns,~T.; Woo,~T.~K.; Farha,~O.; Snurr,~R.~Q.,
  \latin{et~al.}  Inverse Design of Nanoporous Crystalline Reticular Materials
  with Deep Generative Models. \textbf{2020}, \relax
\mciteBstWouldAddEndPunctfalse
\mciteSetBstMidEndSepPunct{\mcitedefaultmidpunct}
{}{\mcitedefaultseppunct}\relax
\EndOfBibitem
\bibitem[Zhou \latin{et~al.}(2019)Zhou, Kearnes, Li, Zare, and
  Riley]{zhou2019optimization}
Zhou,~Z.; Kearnes,~S.; Li,~L.; Zare,~R.~N.; Riley,~P. Optimization of molecules
  via deep reinforcement learning. \emph{Scientific reports} \textbf{2019},
  \emph{9}, 1--10\relax
\mciteBstWouldAddEndPuncttrue
\mciteSetBstMidEndSepPunct{\mcitedefaultmidpunct}
{\mcitedefaultendpunct}{\mcitedefaultseppunct}\relax
\EndOfBibitem
\bibitem[Russo and Golovchenko(2012)Russo, and Golovchenko]{russo2012atom}
Russo,~C.~J.; Golovchenko,~J.~A. Atom-by-atom nucleation and growth of graphene
  nanopores. \emph{Proceedings of the National Academy of Sciences}
  \textbf{2012}, \emph{109}, 5953--5957\relax
\mciteBstWouldAddEndPuncttrue
\mciteSetBstMidEndSepPunct{\mcitedefaultmidpunct}
{\mcitedefaultendpunct}{\mcitedefaultseppunct}\relax
\EndOfBibitem
\bibitem[Feng \latin{et~al.}(2015)Feng, Liu, Graf, Lihter, Bulushev, Dumcenco,
  Alexander, Krasnozhon, Vuletic, Kis, \latin{et~al.}
  others]{feng2015electrochemical}
Feng,~J.; Liu,~K.; Graf,~M.; Lihter,~M.; Bulushev,~R.~D.; Dumcenco,~D.;
  Alexander,~D.~T.; Krasnozhon,~D.; Vuletic,~T.; Kis,~A., \latin{et~al.}
  Electrochemical reaction in single layer MoS2: nanopores opened atom by atom.
  \emph{Nano letters} \textbf{2015}, \emph{15}, 3431--3438\relax
\mciteBstWouldAddEndPuncttrue
\mciteSetBstMidEndSepPunct{\mcitedefaultmidpunct}
{\mcitedefaultendpunct}{\mcitedefaultseppunct}\relax
\EndOfBibitem
\bibitem[LeCun \latin{et~al.}(1998)LeCun, Bottou, Bengio, and
  Haffner]{lecun1998gradient}
LeCun,~Y.; Bottou,~L.; Bengio,~Y.; Haffner,~P. Gradient-based learning applied
  to document recognition. \emph{Proceedings of the IEEE} \textbf{1998},
  \emph{86}, 2278--2324\relax
\mciteBstWouldAddEndPuncttrue
\mciteSetBstMidEndSepPunct{\mcitedefaultmidpunct}
{\mcitedefaultendpunct}{\mcitedefaultseppunct}\relax
\EndOfBibitem
\bibitem[Krizhevsky \latin{et~al.}(2012)Krizhevsky, Sutskever, and
  Hinton]{krizhevsky2012imagenet}
Krizhevsky,~A.; Sutskever,~I.; Hinton,~G.~E. Imagenet classification with deep
  convolutional neural networks. Advances in neural information processing
  systems. 2012; pp 1097--1105\relax
\mciteBstWouldAddEndPuncttrue
\mciteSetBstMidEndSepPunct{\mcitedefaultmidpunct}
{\mcitedefaultendpunct}{\mcitedefaultseppunct}\relax
\EndOfBibitem
\bibitem[Simonyan and Zisserman(2014)Simonyan, and Zisserman]{simonyan2014very}
Simonyan,~K.; Zisserman,~A. Very deep convolutional networks for large-scale
  image recognition. \emph{arXiv preprint arXiv:1409.1556} \textbf{2014},
  \relax
\mciteBstWouldAddEndPunctfalse
\mciteSetBstMidEndSepPunct{\mcitedefaultmidpunct}
{}{\mcitedefaultseppunct}\relax
\EndOfBibitem
\bibitem[He \latin{et~al.}(2016)He, Zhang, Ren, and Sun]{he2016deep}
He,~K.; Zhang,~X.; Ren,~S.; Sun,~J. Deep residual learning for image
  recognition. Proceedings of the IEEE conference on computer vision and
  pattern recognition. 2016; pp 770--778\relax
\mciteBstWouldAddEndPuncttrue
\mciteSetBstMidEndSepPunct{\mcitedefaultmidpunct}
{\mcitedefaultendpunct}{\mcitedefaultseppunct}\relax
\EndOfBibitem
\bibitem[Plimpton(1993)]{plimpton1993fast}
Plimpton,~S. \emph{Fast parallel algorithms for short-range molecular
  dynamics}; 1993\relax
\mciteBstWouldAddEndPuncttrue
\mciteSetBstMidEndSepPunct{\mcitedefaultmidpunct}
{\mcitedefaultendpunct}{\mcitedefaultseppunct}\relax
\EndOfBibitem
\bibitem[Humphrey \latin{et~al.}(1996)Humphrey, Dalke, Schulten, \latin{et~al.}
  others]{humphrey1996vmd}
Humphrey,~W.; Dalke,~A.; Schulten,~K., \latin{et~al.}  VMD: visual molecular
  dynamics. \emph{Journal of molecular graphics} \textbf{1996}, \emph{14},
  33--38\relax
\mciteBstWouldAddEndPuncttrue
\mciteSetBstMidEndSepPunct{\mcitedefaultmidpunct}
{\mcitedefaultendpunct}{\mcitedefaultseppunct}\relax
\EndOfBibitem
\bibitem[Mark and Nilsson(2001)Mark, and Nilsson]{mark2001structure}
Mark,~P.; Nilsson,~L. Structure and dynamics of the TIP3P, SPC, and SPC/E water
  models at 298 K. \emph{The Journal of Physical Chemistry A} \textbf{2001},
  \emph{105}, 9954--9960\relax
\mciteBstWouldAddEndPuncttrue
\mciteSetBstMidEndSepPunct{\mcitedefaultmidpunct}
{\mcitedefaultendpunct}{\mcitedefaultseppunct}\relax
\EndOfBibitem
\bibitem[Ryckaert \latin{et~al.}(1977)Ryckaert, Ciccotti, and
  Berendsen]{ryckaert1977numerical}
Ryckaert,~J.-P.; Ciccotti,~G.; Berendsen,~H.~J. Numerical integration of the
  cartesian equations of motion of a system with constraints: molecular
  dynamics of n-alkanes. \emph{Journal of computational physics} \textbf{1977},
  \emph{23}, 327--341\relax
\mciteBstWouldAddEndPuncttrue
\mciteSetBstMidEndSepPunct{\mcitedefaultmidpunct}
{\mcitedefaultendpunct}{\mcitedefaultseppunct}\relax
\EndOfBibitem
\bibitem[Plimpton \latin{et~al.}(1997)Plimpton, Pollock, and
  Stevens]{plimpton1997particle}
Plimpton,~S.; Pollock,~R.; Stevens,~M. Particle-Mesh Ewald and rRESPA for
  Parallel Molecular Dynamics Simulations. PPSC. 1997\relax
\mciteBstWouldAddEndPuncttrue
\mciteSetBstMidEndSepPunct{\mcitedefaultmidpunct}
{\mcitedefaultendpunct}{\mcitedefaultseppunct}\relax
\EndOfBibitem
\bibitem[Nos{\'e}(1984)]{nose1984unified}
Nos{\'e},~S. A unified formulation of the constant temperature molecular
  dynamics methods. \emph{The Journal of chemical physics} \textbf{1984},
  \emph{81}, 511--519\relax
\mciteBstWouldAddEndPuncttrue
\mciteSetBstMidEndSepPunct{\mcitedefaultmidpunct}
{\mcitedefaultendpunct}{\mcitedefaultseppunct}\relax
\EndOfBibitem
\bibitem[Hoover(1985)]{hoover1985canonical}
Hoover,~W.~G. Canonical dynamics: Equilibrium phase-space distributions.
  \emph{Physical review A} \textbf{1985}, \emph{31}, 1695\relax
\mciteBstWouldAddEndPuncttrue
\mciteSetBstMidEndSepPunct{\mcitedefaultmidpunct}
{\mcitedefaultendpunct}{\mcitedefaultseppunct}\relax
\EndOfBibitem
\bibitem[Perez and Wang(2017)Perez, and Wang]{perez2017effectiveness}
Perez,~L.; Wang,~J. The effectiveness of data augmentation in image
  classification using deep learning. \emph{arXiv preprint arXiv:1712.04621}
  \textbf{2017}, \relax
\mciteBstWouldAddEndPunctfalse
\mciteSetBstMidEndSepPunct{\mcitedefaultmidpunct}
{}{\mcitedefaultseppunct}\relax
\EndOfBibitem
\bibitem[Van~Dyk and Meng(2001)Van~Dyk, and Meng]{van2001art}
Van~Dyk,~D.~A.; Meng,~X.-L. The art of data augmentation. \emph{Journal of
  Computational and Graphical Statistics} \textbf{2001}, \emph{10}, 1--50\relax
\mciteBstWouldAddEndPuncttrue
\mciteSetBstMidEndSepPunct{\mcitedefaultmidpunct}
{\mcitedefaultendpunct}{\mcitedefaultseppunct}\relax
\EndOfBibitem
\bibitem[Larsen \latin{et~al.}(2017)Larsen, Mortensen, Blomqvist, Castelli,
  Christensen, Du{\l}ak, Friis, Groves, Hammer, Hargus, \latin{et~al.}
  others]{larsen2017atomic}
Larsen,~A.~H.; Mortensen,~J.~J.; Blomqvist,~J.; Castelli,~I.~E.;
  Christensen,~R.; Du{\l}ak,~M.; Friis,~J.; Groves,~M.~N.; Hammer,~B.;
  Hargus,~C., \latin{et~al.}  The atomic simulation environment—a Python
  library for working with atoms. \emph{Journal of Physics: Condensed Matter}
  \textbf{2017}, \emph{29}, 273002\relax
\mciteBstWouldAddEndPuncttrue
\mciteSetBstMidEndSepPunct{\mcitedefaultmidpunct}
{\mcitedefaultendpunct}{\mcitedefaultseppunct}\relax
\EndOfBibitem
\bibitem[Nair and Hinton(2010)Nair, and Hinton]{nair2010rectified}
Nair,~V.; Hinton,~G.~E. Rectified linear units improve restricted boltzmann
  machines. ICML. 2010\relax
\mciteBstWouldAddEndPuncttrue
\mciteSetBstMidEndSepPunct{\mcitedefaultmidpunct}
{\mcitedefaultendpunct}{\mcitedefaultseppunct}\relax
\EndOfBibitem
\bibitem[Chen and Guestrin(2016)Chen, and Guestrin]{chen2016xgboost}
Chen,~T.; Guestrin,~C. Xgboost: A scalable tree boosting system. Proceedings of
  the 22nd acm sigkdd international conference on knowledge discovery and data
  mining. 2016; pp 785--794\relax
\mciteBstWouldAddEndPuncttrue
\mciteSetBstMidEndSepPunct{\mcitedefaultmidpunct}
{\mcitedefaultendpunct}{\mcitedefaultseppunct}\relax
\EndOfBibitem
\bibitem[Rogers and Hahn(2010)Rogers, and Hahn]{rogers2010extended}
Rogers,~D.; Hahn,~M. Extended-connectivity fingerprints. \emph{Journal of
  chemical information and modeling} \textbf{2010}, \emph{50}, 742--754\relax
\mciteBstWouldAddEndPuncttrue
\mciteSetBstMidEndSepPunct{\mcitedefaultmidpunct}
{\mcitedefaultendpunct}{\mcitedefaultseppunct}\relax
\EndOfBibitem
\bibitem[Landrum(2016)]{landrum2016rdkit}
Landrum,~G. Rdkit: Open-source cheminformatics software. \emph{GitHub and
  SourceForge} \textbf{2016}, \emph{10}, 3592822\relax
\mciteBstWouldAddEndPuncttrue
\mciteSetBstMidEndSepPunct{\mcitedefaultmidpunct}
{\mcitedefaultendpunct}{\mcitedefaultseppunct}\relax
\EndOfBibitem
\bibitem[Kingma and Ba(2014)Kingma, and Ba]{kingma2014adam}
Kingma,~D.~P.; Ba,~J. Adam: A method for stochastic optimization. \emph{arXiv
  preprint arXiv:1412.6980} \textbf{2014}, \relax
\mciteBstWouldAddEndPunctfalse
\mciteSetBstMidEndSepPunct{\mcitedefaultmidpunct}
{}{\mcitedefaultseppunct}\relax
\EndOfBibitem
\bibitem[Richards(1959)]{richards1959flexible}
Richards,~F. A flexible growth function for empirical use. \emph{Journal of
  experimental Botany} \textbf{1959}, \emph{10}, 290--301\relax
\mciteBstWouldAddEndPuncttrue
\mciteSetBstMidEndSepPunct{\mcitedefaultmidpunct}
{\mcitedefaultendpunct}{\mcitedefaultseppunct}\relax
\EndOfBibitem
\bibitem[Maaten and Hinton(2008)Maaten, and Hinton]{maaten2008visualizing}
Maaten,~L. v.~d.; Hinton,~G. Visualizing data using t-SNE. \emph{Journal of
  machine learning research} \textbf{2008}, \emph{9}, 2579--2605\relax
\mciteBstWouldAddEndPuncttrue
\mciteSetBstMidEndSepPunct{\mcitedefaultmidpunct}
{\mcitedefaultendpunct}{\mcitedefaultseppunct}\relax
\EndOfBibitem
\bibitem[Dulac-Arnold \latin{et~al.}(2015)Dulac-Arnold, Evans, van Hasselt,
  Sunehag, Lillicrap, Hunt, Mann, Weber, Degris, and Coppin]{dulac2015deep}
Dulac-Arnold,~G.; Evans,~R.; van Hasselt,~H.; Sunehag,~P.; Lillicrap,~T.;
  Hunt,~J.; Mann,~T.; Weber,~T.; Degris,~T.; Coppin,~B. Deep reinforcement
  learning in large discrete action spaces. \emph{arXiv preprint
  arXiv:1512.07679} \textbf{2015}, \relax
\mciteBstWouldAddEndPunctfalse
\mciteSetBstMidEndSepPunct{\mcitedefaultmidpunct}
{}{\mcitedefaultseppunct}\relax
\EndOfBibitem
\bibitem[Lalitha \latin{et~al.}(2016)Lalitha, Nataraj, and
  Lakshmipathi]{lalitha2016calcium}
Lalitha,~M.; Nataraj,~Y.; Lakshmipathi,~S. Calcium decorated and doped
  phosphorene for gas adsorption. \emph{Applied Surface Science} \textbf{2016},
  \emph{377}, 311--323\relax
\mciteBstWouldAddEndPuncttrue
\mciteSetBstMidEndSepPunct{\mcitedefaultmidpunct}
{\mcitedefaultendpunct}{\mcitedefaultseppunct}\relax
\EndOfBibitem
\bibitem[Sheberla \latin{et~al.}(2017)Sheberla, Bachman, Elias, Sun, Shao-Horn,
  and Dinc{\u{a}}]{sheberla2017conductive}
Sheberla,~D.; Bachman,~J.~C.; Elias,~J.~S.; Sun,~C.-J.; Shao-Horn,~Y.;
  Dinc{\u{a}},~M. Conductive MOF electrodes for stable supercapacitors with
  high areal capacitance. \emph{Nature materials} \textbf{2017}, \emph{16},
  220--224\relax
\mciteBstWouldAddEndPuncttrue
\mciteSetBstMidEndSepPunct{\mcitedefaultmidpunct}
{\mcitedefaultendpunct}{\mcitedefaultseppunct}\relax
\EndOfBibitem
\bibitem[Barati~Farimani and Aluru(2011)Barati~Farimani, and
  Aluru]{barati2011spatial}
Barati~Farimani,~A.; Aluru,~N.~R. Spatial diffusion of water in carbon
  nanotubes: from fickian to ballistic motion. \emph{The Journal of Physical
  Chemistry B} \textbf{2011}, \emph{115}, 12145--12149\relax
\mciteBstWouldAddEndPuncttrue
\mciteSetBstMidEndSepPunct{\mcitedefaultmidpunct}
{\mcitedefaultendpunct}{\mcitedefaultseppunct}\relax
\EndOfBibitem
\bibitem[Joung and Cheatham~III(2008)Joung, and
  Cheatham~III]{joung2008determination}
Joung,~I.~S.; Cheatham~III,~T.~E. Determination of alkali and halide monovalent
  ion parameters for use in explicitly solvated biomolecular simulations.
  \emph{The journal of physical chemistry B} \textbf{2008}, \emph{112},
  9020--9041\relax
\mciteBstWouldAddEndPuncttrue
\mciteSetBstMidEndSepPunct{\mcitedefaultmidpunct}
{\mcitedefaultendpunct}{\mcitedefaultseppunct}\relax
\EndOfBibitem
\bibitem[Bradski(2000)]{opencv_library}
Bradski,~G. {The OpenCV Library}. \emph{Dr. Dobb's Journal of Software Tools}
  \textbf{2000}, \relax
\mciteBstWouldAddEndPunctfalse
\mciteSetBstMidEndSepPunct{\mcitedefaultmidpunct}
{}{\mcitedefaultseppunct}\relax
\EndOfBibitem
\end{mcitethebibliography}

\newpage

\section{Lennard-Jones potentials for MD simulations}

The 12-6 Lennard-Jones (LJ) potentials used in MD simulations for generating dataset are tabulated as below. Force field between different types of atoms are calculated using Lorentz-Berthelot rule.

\begin{table}
\renewcommand{\thetable}{S1}
  \caption{12-6 Lennard-Jones potentials}
  \label{tb:lj}
  \begin{tabular}{l@{\hspace{2cm}}l@{\hspace{2cm}}l}
    \hline
    Interaction & $\sigma$ ($\si{\angstrom}$) & $\epsilon$ (kcal mol$^{-1}$)  \\
    \hline
    C-C\cite{barati2011spatial}  & 3.3900 & 0.0692 \\
    O-O\cite{barati2011spatial}  & 3.1656 & 0.1554 \\
    H-H\cite{barati2011spatial}  & 0.0000 & 0.0000  \\
    K-K\cite{joung2008determination}  & 2.8384 & 0.4297\\
    Cl-Cl\cite{joung2008determination}  & 4.8305 & 0.0128\\
    \hline
  \end{tabular}
\end{table}

\section{Details of XGBoost Regression Model}

For XGBoost \cite{chen2016xgboost}, 100 iterations of random search with 5-fold cross validation was applied on the hyperparameter grid tabulated as below. The hyperparameters of best XGBoost models for both water flux and ion rejection rate prediction are also recorded. 

\begin{table}
  \renewcommand{\thetable}{S2}
  \caption{Hyperparameter grid}
  \label{tb:hypergrid}
  \begin{tabular}{l@{\hspace{2cm}}l}
    \hline
    Hyperparameter & Value  \\
    \hline
    learning\_rate $\alpha$  & 0.01, 0.05, 0.1, 0.15, 0.2, 0.25, 0.3 \\
    max\_depth  & 3, 5, 7, 9, 10, 12, 15 \\
    min\_child\_weight  & 1, 2, 3, 4, 5, 6 \\
    gamma $\gamma$  & 0.0, 0.1, 0.2, 0.3, 0.4\\
    colsample\_bytree  & 0.3, 0.5, 0.7, 0.9, 1, 1.2\\
    \hline
  \end{tabular}
\end{table}

\begin{table}
  \renewcommand{\thetable}{S3}
  \caption{Best hyperparameters for XGBoost models}
  \label{tb:besthyper}
  \begin{tabular}{l@{\hspace{2cm}}l@{\hspace{2cm}}l}
    \hline
    Best hyperparameter & Water flux model & Ion rejection model  \\
    \hline
    learning\_rate $\alpha$  & 0.2 & 0.1\\
    max\_depth  & 7 & 12\\
    min\_child\_weight  & 1 & 5\\
    gamma $\gamma$  & 0.0 & 0.0\\
    colsample\_bytree  & 0.3 & 1\\
    \hline
  \end{tabular}
\end{table}

\section{Details of DRL agent}

\begin{figure}[htb!]
\renewcommand{\thefigure}{S1}
    \centering
    \includegraphics[width=.85\linewidth]{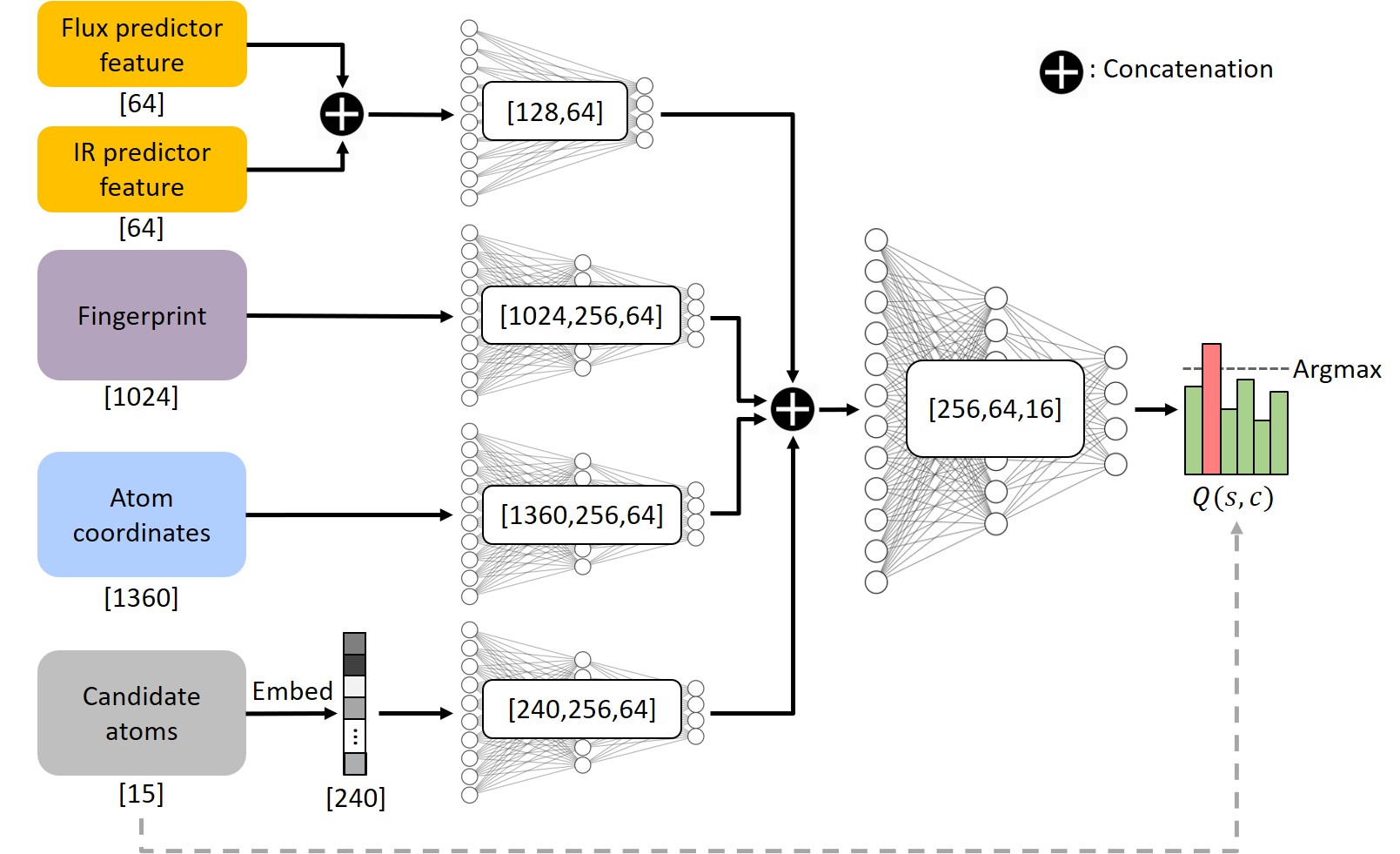}
    \caption{Overview of deep Q network architecture. The numbers below each box illustrates the dimensions.}
    \label{fig:dqn}
\end{figure}

Deep Q network \cite{mnih2013playing,mnih2015human} is utilized to model the DRL agent, which takes as input the graphene state representation and candidate atoms(Fig.~\ref{fig:dqn}). The state representation is consisted of three components: graphene geometrical features, Morgan finger fingerprint\cite{rogers2010extended} and atom coordinates. The geometrical features are extracted from both the water flux and ion rejection prediction CNN models\cite{simonyan2014very,he2016deep}. All the three components are mapped to 64-dimensional vectors via MLPs. Besides, the candidate atoms are represented with a 608-dimensional vector where each element corresponds one removable atom. Each entry of the vector is set to one if the corresponding atom is a candidate and zero elsewhere, and is mapped to a 16-dimensional representation through an embedding layer. The embedding is then mapped to a 64-dimensional vector and fed into a MLP in combination of state representations to predict the Q values of all the candidate actions. During test time, the action with the maximum Q value is executed to interact with the graphene. However, to encourage exploration of the DRL agent during training, the action is selected based on $\epsilon$-greedy \cite{sutton1998introduction}. Namely, with probability $\epsilon$, the agent will randomly pick an action from the candidates. In our setting, $\epsilon$ is initially set to $0.9$ and decayed to $0.05$ linearly after 1000 episodes. 

\begin{figure}[!htb]
\renewcommand{\thefigure}{S2}
    \begin{subfigure}{0.49\textwidth}
      \centering
      \includegraphics[width=\linewidth]{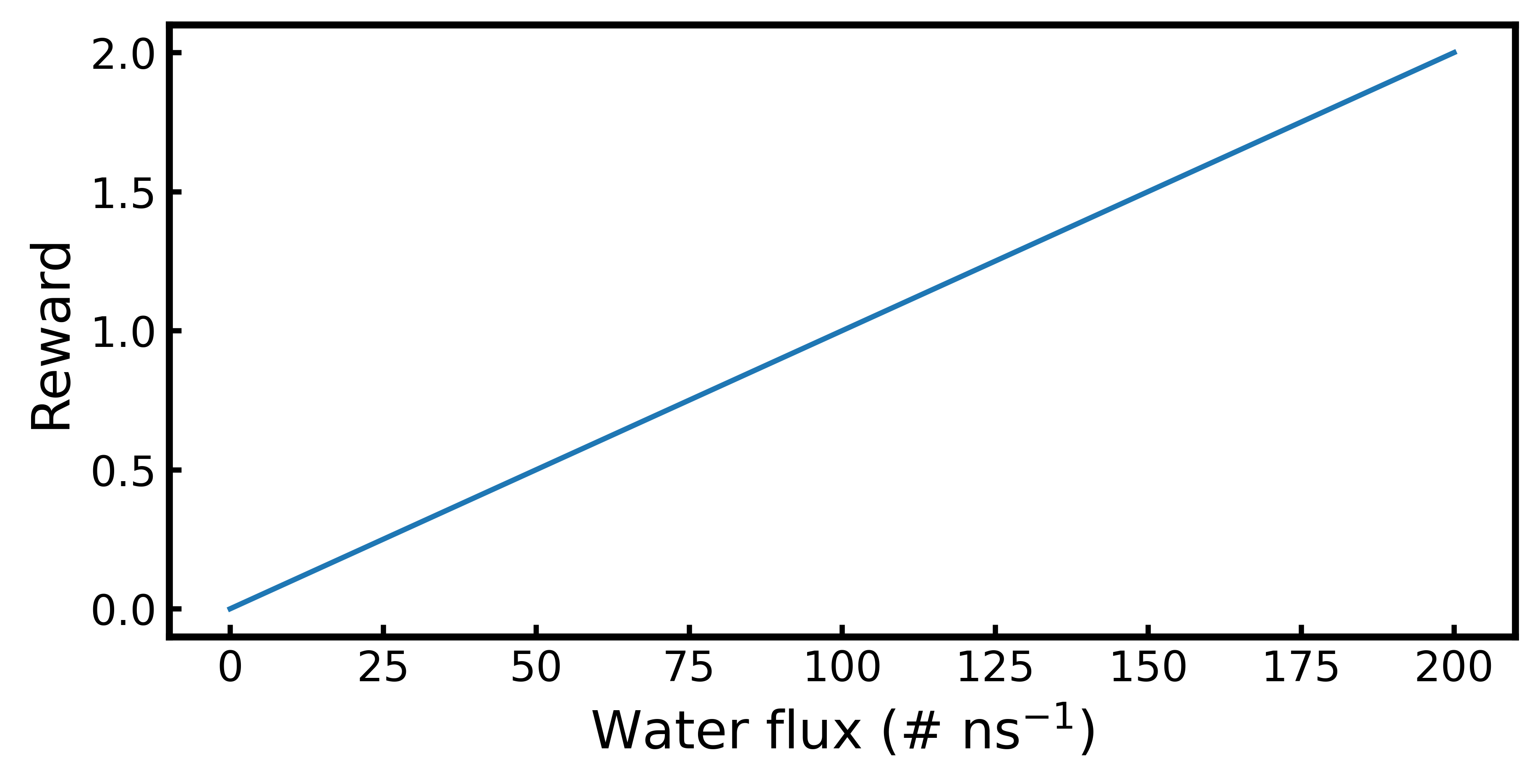}
      \caption{}
      \label{fig:flux_rew}
    \end{subfigure}
    \hfill
    \begin{subfigure}{0.49\textwidth}
      \centering
      \includegraphics[width=\linewidth]{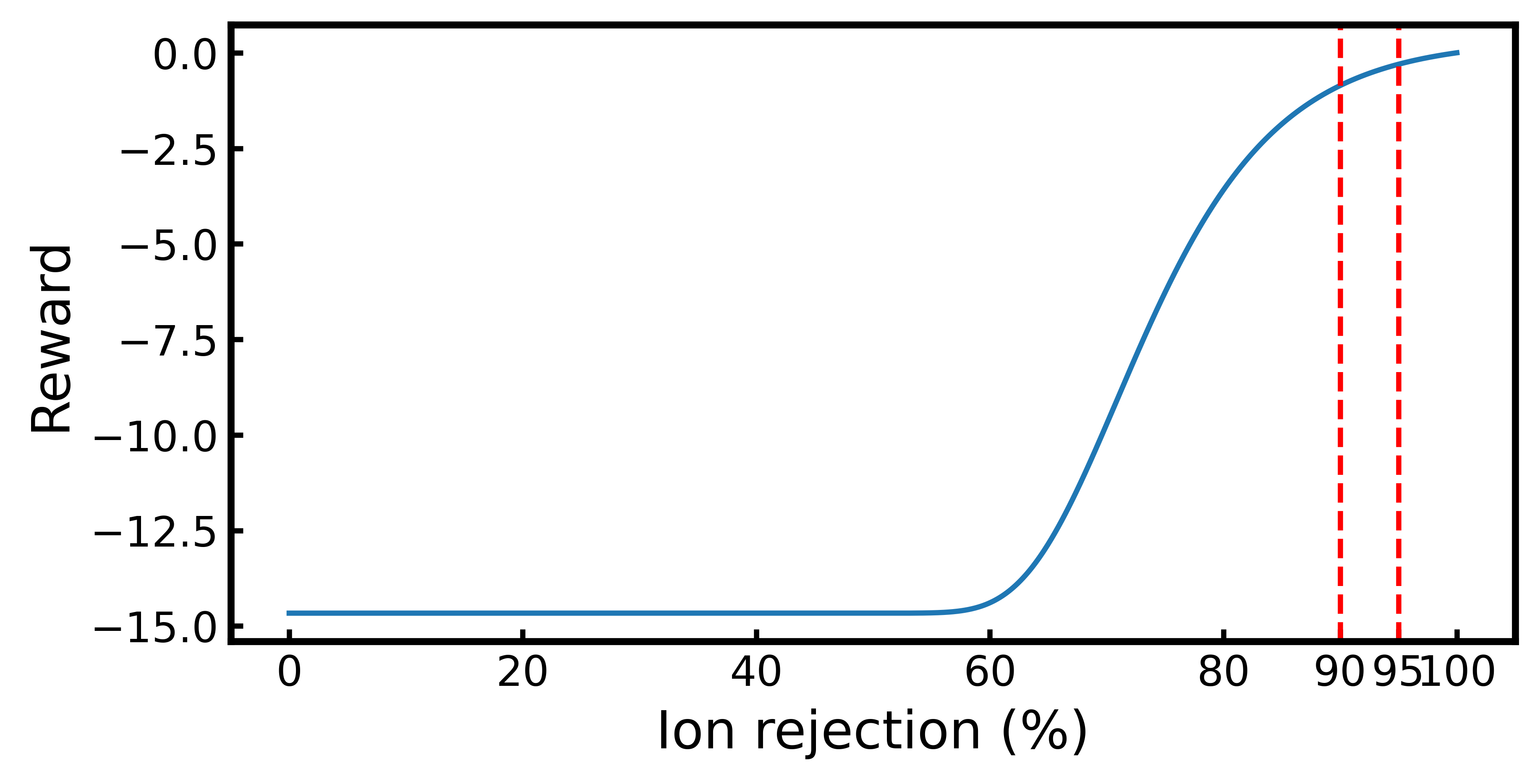}
      \caption{}
      \label{fig:rej_rew}
    \end{subfigure}
\caption{(a) Water flux vs. corresponding reward value.  (b) Ion rejection vs. corresponding reward value}
\label{fig:flux_rej_rew}
\end{figure}

Also, the generalized logistic function \cite{richards1959flexible} utilized in the reward function for penalizing ion rejection is illustrated in Fig.~\ref{fig:flux_rej_rew}. When the ion rejection is greater than $95\%$, the function returns a near-zero value. Therefore, given that DRL agent receives positive reward for removing atoms and generating water flux, it tends to aggressively remove atoms at the beginning of each episode. While the ion rejection drops below $90\%$, the function gives a harsh penalty to the agent. Such reward function strongly discourage the agent from creating a nanoporus graphene with low ion rejection rate, which is inefficient in real-world water desalination.

\section{Comparison between original and augmented pores}
MD simulations are run with a human-designed pore and its flip-translate augmented version. Each data point is calculated by averaging water flux and ion rejection rate of 4 simulations. Error bar represent the value of one standard deviation. The water flux difference between the original pore and augmented pore is 3.17 \#/ns (approximately 2\%) and the ion rejection rate difference is 0.6\%. All differences are within the error bar of the result of 4 simulations. From this result, we can conclude that the augmentation method implemented in this work is valid.
\begin{figure}[]
\renewcommand{\thefigure}{S3}
    \centering
    \includegraphics[width=.6\linewidth]{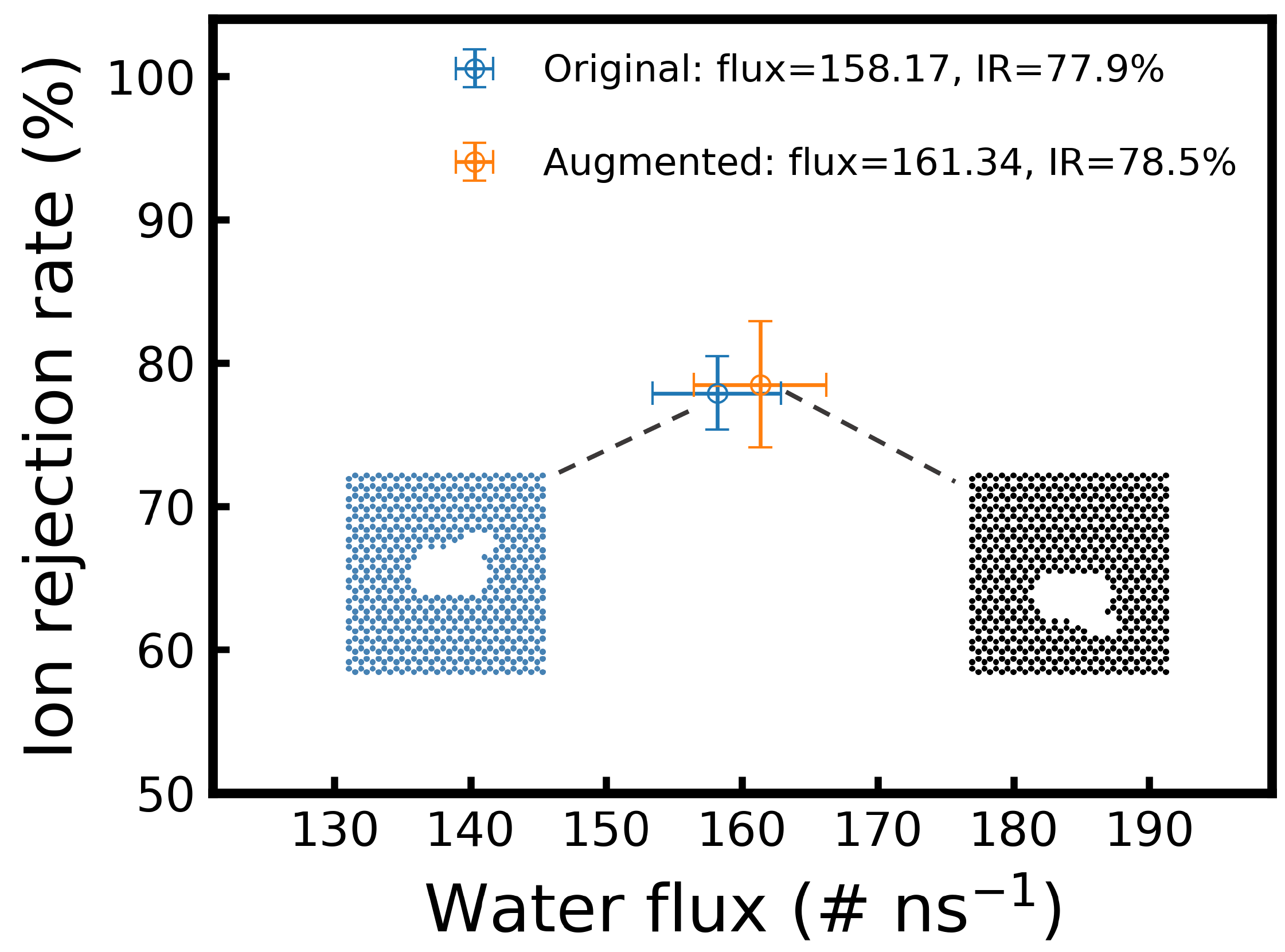}
    \caption{Comparison between the water desalination performance of original human-designed graphene nanopore (left, blue membrane) and its augmented copy version (right, black membrane). The performance difference between them is minimum and within the error bar.}
    \label{fig:ori_vs_aug}
\end{figure}

\section{Pore area calculation}
\begin{figure}[]
\renewcommand{\thefigure}{S4}
    \begin{subfigure}{0.49\textwidth}
      \centering
      \includegraphics[width=0.7\linewidth]{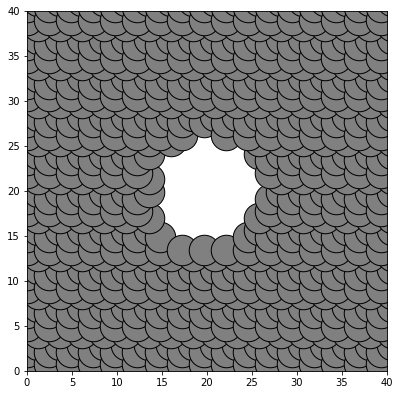}
      \caption{}
      \label{fig:cv_pore}
    \end{subfigure}
    \hfill
    \begin{subfigure}{0.49\textwidth}
      \centering
      \includegraphics[width=0.7\linewidth]{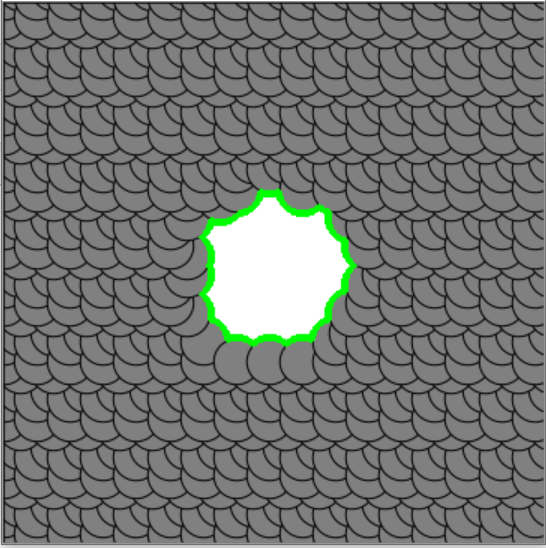}
      \caption{}
      \label{fig:cv_pore_detect}
    \end{subfigure}
\caption{(a) Plot nanoporous graphene membrane  (b) Detecting pore area and calculate its size using computer vision method}
\label{fig:calcArea}
\end{figure}

The area of graphene nanopores mentioned in this work is calculated using computer vision method (OpenCV package\cite{opencv_library}). For each nanoporous graphene membrane, all of its atoms which locate in the area of $0\si{\angstrom} \leq x \leq 40\si{\angstrom}$ and $0\si{\angstrom} \leq y \leq 40\si{\angstrom}$ are plotted with radius as $\sigma$ (3.39$\si{\angstrom}$ used in this paper) of carbon (Fig.~\ref{fig:cv_pore}). The graphene membrane image dimension in pixel is 360 $\times$ 360. The OpenCV package is then used to detect the pore from the image (Fig.~\ref{fig:cv_pore_detect}), and the area of the pore in terms of pixel is calculated. Further, The area of pore in terms of $\si{\angstrom}^2$ can be calculated using the ratio: 1 pixel = $(40/360)^2$ $\si{\angstrom}^2$.

\section{Compilation of DRL generated pores with high predicted performance}
\begin{figure}[]
\renewcommand{\thefigure}{S5}
    \centering
    \includegraphics[width=0.7\linewidth]{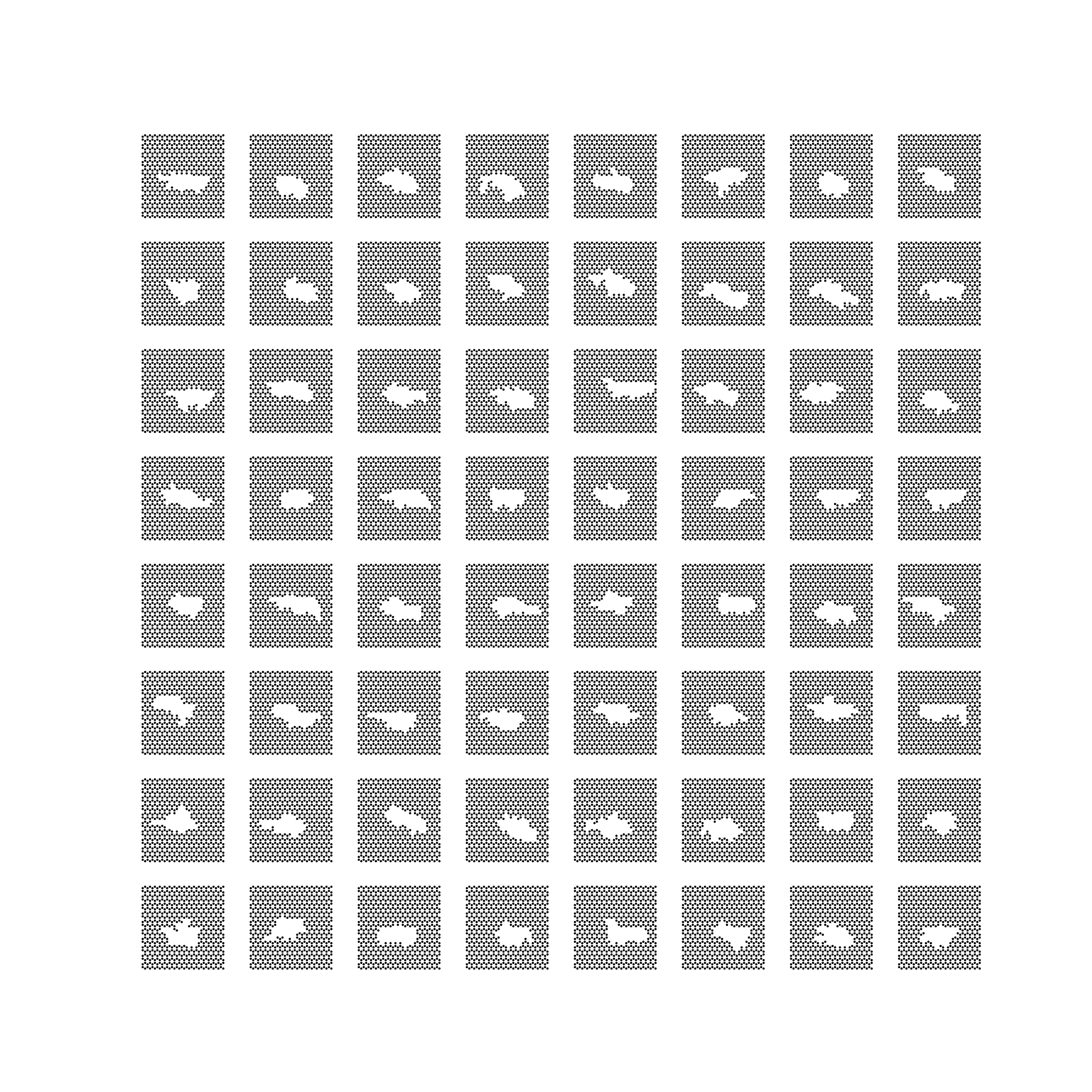}
\caption{DRL generated pores with predicted water flux $>$ 120 \#/ns and ion rejection rate $>$ 90\%.}
\label{fig:compilation}
\end{figure}

The compilation of 64 DRL generated pores with predicted water flux $>$ 120 \#/ns and ion rejection rate $>$ 90\% are demonstrated in (Fig.~\ref{fig:compilation}). One common trait of those grpahene nanopores is that they tend to have narrow cavities. Those narrow cavities in pore allow the passage of water molecules but are small enough to restrict the transport of ions along with their hydration shell. DRL gradually learned the importance of narrow cavity and applied it during the optimization of graphene nanopores.

\end{document}